%% file: main.tex
\documentclass{article}

    \PassOptionsToPackage{numbers, compress}{natbib}

\usepackage[preprint]{neurips_2025}

\usepackage[pdftex]{graphicx}

\usepackage[utf8]{inputenc} 
\usepackage[T1]{fontenc}    
\usepackage{hyperref}       
\usepackage{url}            
\usepackage{booktabs}       
\usepackage{amsfonts}       
\usepackage{nicefrac}       
\usepackage{microtype}      
\usepackage{xcolor}         
\usepackage{graphicx}
\usepackage{float}
\usepackage{amsmath}
\usepackage{amsthm}
\usepackage{xspace}
\usepackage{cleveref}
\usepackage{subcaption}
\usepackage{booktabs}
\usepackage{bbding} 
\usepackage{pifont}
\usepackage{diagbox}
\usepackage{multirow}

\usepackage[linesnumbered,ruled,vlined]{algorithm2e}
\RestyleAlgo{ruled}

\usepackage[symbol]{footmisc}

\newcommand{\cmark}{\ding{51}} 
\newcommand{\xmark}{\ding{55}} 

\theoremstyle{remark}

\theoremstyle{definition}
\newtheorem{definition}{Definition}

\newcommand{\ttti}[1]{\texttt{i}\xspace}

\usepackage{wrapfig}

\usepackage{enumitem}

\newcommand{\benchmarkname}{T-GRAB\xspace}

\input{macros}

\title{T-GRAB: A Synthetic Diagnostic Benchmark for Learning on Temporal Graphs }

\input{tex/000authors}

\begin{document}

\maketitle

\input{tex/000abstract}

\input{tex/010intro}
\input{tex/020related}

\input{tex/030prelims}
\input{tex/040method}
\input{tex/050periodic}

\input{tex/060cause}
\input{tex/070long}
\input{tex/080discussion}

\input{tex/085conclusion}

\bibliographystyle{plainnat}
\bibliography{ref.bib}

\clearpage
\newpage
\appendix
\input{tex/100appendix}

\end{document}

%% file: macros.tex
\newcommand{\mat}[1]{\ensuremath{\mathbf{#1}}}

\newcommand{\lag}{\ell}

\newcommand{\Pdet}[1]{\mathcal{P}^\text{det}(#1)}
\newcommand{\Psto}[1]{\mathcal{P}^\text{sto}(#1)}
\newcommand{\causeeffect}[1]{ \mathcal{C}\hspace{-0.03cm}\mathcal{E}(#1)}
\newcommand{\longrange}[1]{ \mathcal{L}\hspace{-0.035cm}\mathcal{R}(#1)}

\newcommand{\first}[1]{\textbf{\textcolor{red}{#1}}}
\newcommand{\second}[1]{\underline{\textcolor{blue}{#1}}}
\newcommand{\third}[1]{\emph{\textcolor{violet}{#1}}}

\SetKwComment{Comment}{//}{}
\SetKwInput{KwInput}{Input}
\SetKwInput{KwParam}{Param}

%% file: tex/000authors.tex
\author{%
Alireza Dizaji\textsuperscript{1,2}\quad Benedict Aaron Tjandra\textsuperscript{1,3}\quad Mehrab Hamidi\textsuperscript{1,2} \\
\textbf{Shenyang Huang}\textsuperscript{1,3,4} \quad \textbf{Guillaume Rabusseau}\textsuperscript{1,2,5}\\
\textsuperscript{1}Mila, 
\textsuperscript{2}DIRO-UdeM,
\textsuperscript{3}University of Oxford,
\textsuperscript{4}SoCS-McGill,
\textsuperscript{5}CIFAR AI Chair
  \vspace*{-0.25cm}
}

%% file: tex/000abstract.tex
\begin{abstract}

Dynamic graph learning methods have recently emerged as powerful tools for modelling relational data evolving through time. However, despite extensive benchmarking efforts, it remains unclear whether current Temporal Graph Neural Networks (TGNNs) effectively capture core temporal patterns such as periodicity, cause-and-effect, and long-range dependencies. In this work, we introduce the\textbf{ Temporal Graph Reasoning Benchmark (\benchmarkname)}, a comprehensive set of synthetic tasks designed to systematically probe the capabilities of TGNNs to reason across time. \benchmarkname provides controlled, interpretable tasks that isolate key temporal skills: counting/memorizing periodic repetitions, inferring delayed causal effects, and capturing long-range dependencies over both spatial and temporal dimensions. We evaluate 11 temporal graph learning methods on these tasks, revealing fundamental shortcomings in their ability to generalize temporal patterns. Our findings offer actionable insights into the limitations of current models, highlight challenges hidden by traditional real-world benchmarks, and motivate the development of architectures with stronger temporal reasoning abilities. The code for T-GRAB can be found at: \url{https://github.com/alirezadizaji/T-GRAB}.
\end{abstract}

%% file: tex/010intro.tex

\section{Introduction}
\looseness-1 Many real-world networks, such as social media networks~\cite{tgb}, human contact networks~\cite{shirzadkhani2024static} and financial transaction~\cite{shamsi2024graphpulse} networks can be formulated as temporal graphs or graphs that evolve over time. Recently, Temporal Graph Neural Networks (TGNNs) have emerged as promising architectures to address the unique challenges associated with ML on temporal graphs, which necessitates the modeling of both spatial and temporal dependencies~\citep{tgn, tgat, dygformer, skarding_tg_survey, ctan, feng_tg_survey}. Naturally, the development of TGNNs is quickly followed by an increased focus to design challenging benchmarks to understand their capabilities~\citep{poursafaei2022betterevaluationdynamiclink, tgb, tgb2.0, tgb-seq, shirzadkhani2024towards} across node, edge, and graph-level tasks. These benchmarks provide significant challenges for TGNNs in both scale and domain diversity with a focus on real-world tasks. However, current TGNNs have been shown to significantly struggle in these benchmarks and, in some cases, even underperform simple heuristics such as EdgeBank~\citep{poursafaei2022betterevaluationdynamiclink} and persistent forecast~\citep{tgb}. 

\looseness-1 When compared to the increasing number of novel architectures proposed, exploring the weaknesses of TGNNs remains under-explored and often applies only to specific categories of methods~ \citep{pint, weisfeiler-go-dynamic, tjandra2024enhancingexpressivitytemporalgraph}. Therefore, we argue that there is a strong need for a surgical and well-designed benchmark to highlight the weakness of existing models in performing crucial yet simple tasks on temporal graphs.

\looseness-1 In the past, diagnostic benchmarks were developed with different task classes to provide crucial insights into model capabilities precisely when complex, real-world benchmarks proved insufficient for pinpointing specific failure modes. For instance, in computer vision, the CLEVR dataset \citep{clevr} utilized synthetically generated scenes to test the compositional reasoning abilities of visual question-answering models, revealing limitations obscured by the biases and confounding factors present in natural images. Similarly, in early natural language processing, the bAbI dataset \citep{babi} provided a suite of 20 synthetic question-answering tasks generated algorithmically to probe basic reasoning skills (such as counting, induction, and deduction) essential for language understanding, offering metrics for progress on these core competencies. Reinforcement learning has also benefited from such focused evaluations; for instance, the Behaviour Suite for RL \citep{bsuite} includes controlled environments specifically designed to diagnose the memory and exploration capabilities of RL agents. 

\looseness-1 These examples demonstrate the power of purely synthetic datasets and environments designed for diagnostic evaluation: they allow for precise control over task complexity and the factors being tested, yielding clear insights into model strengths and weaknesses. Such a dedicated, synthetic diagnostic benchmark is currently missing for the domain of temporal graph learning (TGL). While existing benchmarks effectively test performance on complex, real-world dynamics, they inherently entangle various challenges – noisy interactions, complex graph structures evolving simultaneously with temporal patterns, and diverse event types. This makes it difficult to determine if a model's failure stems from an inability to handle graph complexity or from a fundamental deficit in capturing specific temporal patterns like periodicity, cause-and-effect relationships, or dependencies spanning long time horizons.

\begin{figure}[!t]
\vspace*{-0.5cm}
    \centering
    \includegraphics[width=0.75\textwidth, keepaspectratio]{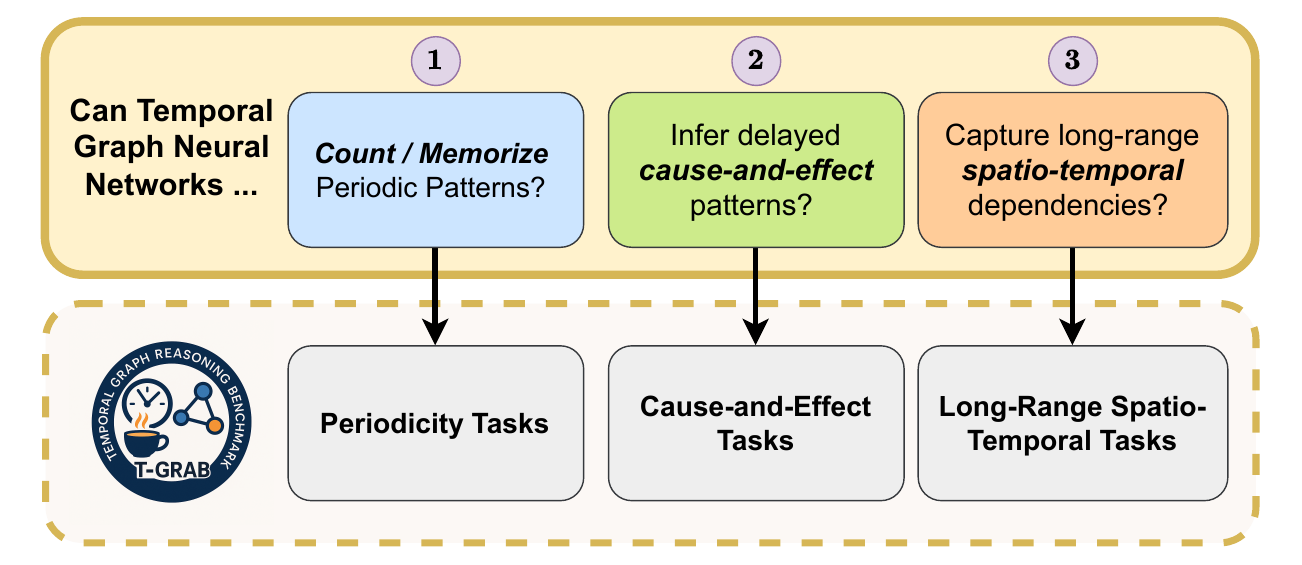}
    \caption{\benchmarkname tests the capabilities of TGNNs to reason over time in three fundamental aspects and includes three carefully designed tasks.}
    \label{fig:tgrab}
    \vspace{-10pt}
\end{figure}

\looseness-1 To address this gap and facilitate a deeper, more interpretable understanding of TGNN limitations, we introduce the Temporal Graph Reasoning Benchmark (\benchmarkname). \benchmarkname (\Cref{fig:tgrab}) comprises a suite of purely synthetic, dynamic graph datasets explicitly designed to probe the fundamental temporal reasoning capabilities essential for modeling real-world dynamic systems. By isolating core temporal patterns within controlled graph structures, \benchmarkname allows for a clear assessment of how well current TGNNs capture and generalize these patterns.


\textbf{Contributions.} Our main contributions are as follows:
\begin{itemize}[topsep=0pt] \itemsep0em 
    \item We introduce \benchmarkname, the first synthetic benchmark designed to systematically evaluate temporal reasoning capabilities of TGNNs in controlled settings. It features three carefully crafted dynamic link prediction tasks: 1) \textbf{periodicity} to assess \textit{temporal pattern counting and memorization}, 2) \textbf{cause-and-effect} to evaluate \textit{delayed dependency inference}, and 3) \textbf{long-range spatio-temporal} to test \textit{long-range spatial-temporal dependency modeling}.
    
    \item \benchmarkname provides a configurable environment where task difficulty can be precisely adjusted to identify the limitations of TGNNs. Our experiments reveal distinct behavioral patterns between CTDG and DTDG methods, and highlight that the number of temporal neighbors, an often overlooked hyperparameter, significantly impacts model performance across our benchmark tasks.
    

     \item For the periodicity tasks, GC-LSTM consistently performs best, even in the most challenging settings, indicating its recurrent structure is better suited for capturing periodic patterns and counting. In the cause-and-effect task, all models struggle with long-term memory, though DyGFormer, TGAT, and TGN degrade most gracefully. Finally, in the spatio-temporal task, DyGFormer's transformer-based architecture excels with short-range spatial dependencies, while TGAT and TGN outperform it as spatial dependencies grow longer.

    \item Notably, no single model consistently outperforms across all tasks in \benchmarkname, contrasting with real-world benchmarks where leaderboards are typically dominated by a few methods. This finding underscores the value of \benchmarkname as a diagnostic tool that can guide the development of more robust and versatile temporal graph learning methods capable of handling diverse temporal reasoning challenges.
\end{itemize}

%% file: tex/020related.tex
\textbf{Related Work}
Temporal Graph Neural Networks (TGNNs) are categorized into continuous and discrete time dynamic graph methods (CTDG and DTDG). CTDG methods process event streams with timestamps using neighbor sampling to model evolving relations. They include TGAT \citep{tgat} with self-attention and temporal encoding, TGN \citep{tgn} with memory modules, DyGFormer \citep{dygformer} using multi-head attention on temporal patches, and CTAN \citep{ctan} employing ODEs. DTDG methods operate on regularly-spaced graph snapshots and typically use recurrent neural networks to track history; popular methods include GC-LSTM \citep{gclstm}, T-GCN \citep{tgcn}, and EvolveGCN \citep{evolvegcn}). 
Recently, \cite{huangutg} compared CTDG and DTDG methods and found that DTDG methods sacrifice accuracy for efficiency by computing on coarser-level snapshot information. Subsequently, they showed that DTDG methods can operate on CTDG datasets by using the Unified Temporal Graph (UTG) framework. In \Cref{sec:tgrab}, we evaluate and compare the above methods from both continuous and discrete-time approaches on \benchmarkname datasets to analyze their capabilities in capturing core temporal patterns. Dedicated benchmarks \citep{Jodie_Kumar_2019, uci, enron, social-evo} have significantly improved TGNN evaluation, though \cite{poursafaei2022betterevaluationdynamiclink} noted overly optimistic results due to simple negative edges and introduced new sampling strategies and diverse datasets. Subsequently, TGB \citep{tgb} and TGB 2.0 \citep{tgb2.0} presented larger, challenging datasets for link and node prediction, while TGB-Seq \citep{tgb-seq} proposed real-world  with complex sequential dynamics and low edge repeatability, revealing limitations in current TGNN generalization. Our work complements these recent efforts by constructing synthetic tasks to more exactly pinpoint the current functional weaknesses of TGNNs.

%% file: tex/030prelims.tex
\section{Temporal Graph Preliminaries}

\begin{definition}[Discrete Time Dynamic Graphs] 
A \emph{Discrete Time Dynamic Graph} $\mathcal{G}$ is a sequence of graph snapshots sampled at regularly-spaced time intervals~\cite{kazemi2020representation}:
$ \mathcal G = G_1,G_2,G_3,\cdots,G_T.$
Each $G_t = (V_t, E_t, \mat X_t)$ is the graph at snapshot $t=1,\cdots, T$, where ${V}_t$, ${E}_t$ are the set of nodes and edges in ${G}_t$, respectively, and $\mat X_t\in \mathbb R^{|V_t| \times d}$ is the matrix of node features at time $t$.
\end{definition}

In this work, we focus on tasks designed for Discrete Time Dynamic Graphs~(DTDGs). As Continuous Time Dynamic Graph~(CTDG) methods can be applied on DTDGs as well~\cite{huangutg}, we benchmark the performance of both types of methods for comprehensive evaluation.
In our tasks, the set of vertices is the same at all time steps, i.e., $V_1=V_2=\cdots =V_t = \cdots$. Node features are also constant through time and consists of one-hot encoding of the $N$ nodes~(i.e., $d=N$ and $\mat X_t = \mat I$ is the identity matrix for all $t$). While dynamic graphs sometimes have edge features, we do not use any in \benchmarkname. We follow the methodology outlined in~\cite{huangutg} to evaluate CTDG methods on discrete time graphs by translating all edges in each graph snapshot $G_t$ into a batch of edges $\{(u,v,t) \mid (u,v)\in E_t\}$. 

%% file: tex/040method.tex
\section{Temporal Graph Reasoning Benchmark: \benchmarkname \raisebox{-1mm}{\includegraphics[height=1.5em]{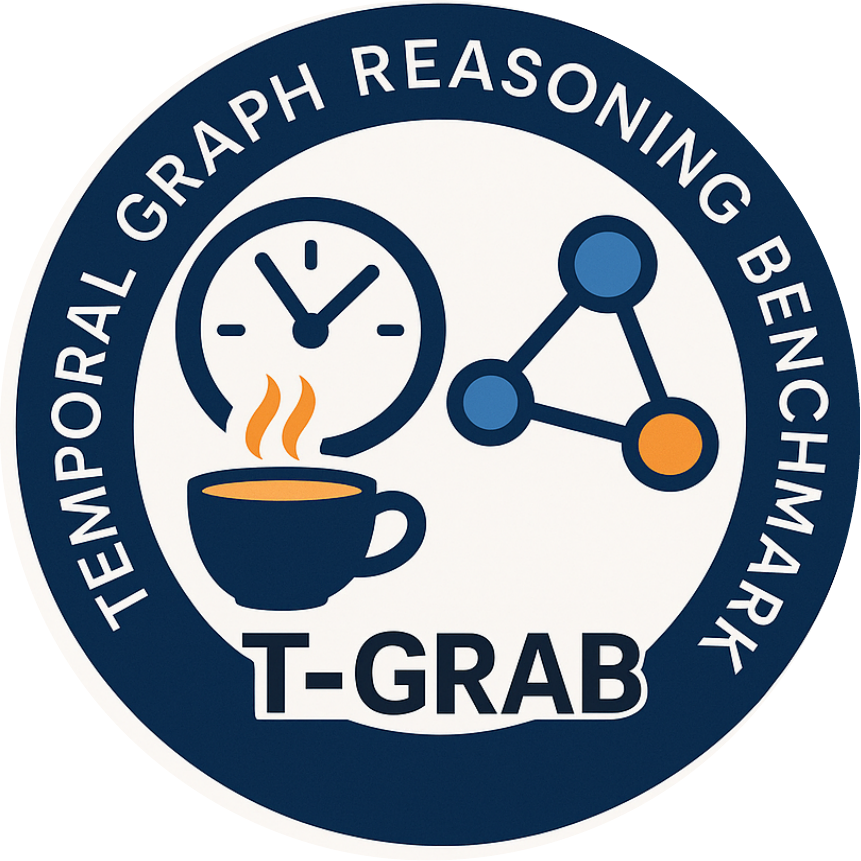}
 }}\label{sec:tgrab}

\looseness-1 In this section, we introduce the Temporal Graph Reasoning Benchmark~(\benchmarkname), the first synthetic benchmark designed to systematically evaluate the temporal reasoning capabilities of TGNNs in a controlled environment. \benchmarkname comprises three categories of dynamic link prediction tasks, each probing distinct aspects of temporal reasoning: 1) \emph{periodicity} tasks, which assess counting and memorization capabilities; 2) \emph{cause-and-effect} tasks, which evaluate the ability to identify causal relationships across time delays; and 3) \emph{long-range spatio-temporal} tasks, which measure how effectively models capture dependencies spanning both spatial and temporal dimensions. These task families are examined in detail in Sections~\ref{synth-benchmark:periodicity}, ~\ref{synth-benchmark:cause}, and ~\ref{synth-benchmark:long}, respectively. As summarized in \Cref{tab:data}, \benchmarkname encompasses a diverse spectrum of graph characteristics and temporal patterns, providing a comprehensive framework to rigorously test the fundamental reasoning capabilities of TGNNs.

\begin{table*}[t]
\caption{\benchmarkname dataset statistics and characteristics. } \label{tab:data} 
  \resizebox{\linewidth}{!}{%
  \begin{tabular}{ l | cccccc}
  \toprule
  Dataset & \# Nodes & \# Edges
  & \# Timestamp & Counting & Memorizing & \begin{tabular}{cc}
       Spatial  \\ Understanding
  \end{tabular}  \\ 
  \midrule
  Periodicity &  100 & 10,560 - 9,144,440 & 96 - 12,288 & \cmark & \cmark & \xmark\\ 
  Cause-and-Effect &  101 & 164,856 - 174,470 & 4,001 - 4,256 & \xmark & \cmark & \xmark\\ 
  Long-Range Spatio-Temporal &  102 & 48,006 - 411,072 & 4,001 - 4,032 & \xmark & \cmark & \cmark\\ 
  \bottomrule
  \end{tabular}
  }
\end{table*}


\looseness-1 \textbf{Methods in Comparison.} We conduct a comprehensive evaluation of diverse Temporal Graph Learning (TGL) approaches on \benchmarkname tasks (experimental details in \Cref{appendix:experimental-details}). Our analysis encompasses four continuous-time (CTDG) architectures (DyGFormer~\cite{dygformer}, CTAN~\cite{ctan}, TGN~\cite{tgn}, and TGAT~\cite{tgat}), three discrete-time (DTDG) frameworks (EvolveGCN~\cite{evolvegcn}, T-GCN~\cite{tgcn}, and GC-LSTM~\cite{gclstm}), two static graph methods (GCN~\citep{gcn} and GAT~\citep{gat}), for which we use their DTDG implementations provided by UTG~\citep{huangutg}, and two established baselines (the \textit{persistence} heuristic, which predicts edges from the previous timestep, and $\text{Edgebank}_{\infty}$~\citep{poursafaei2022betterevaluationdynamiclink}). Model sizes are detailed in Appendix~\ref{appendix:parameters}. This selection represents the state-of-the-art across different temporal graph learning paradigms, enabling a rigorous assessment of their fundamental reasoning capabilities. 


\textbf{Evaluation Protocols.}
Prior research has demonstrated that evaluation results for dynamic link prediction can vary substantially depending on negative sampling strategies~\cite{poursafaei2022betterevaluationdynamiclink}. To ensure methodological rigor and reproducibility, we implement a comprehensive evaluation framework that calculates the average $F_1$ score across all possible node pairs at each test time step. This approach eliminates sampling bias and provides a more reliable performance assessment. For periodicity tasks, our evaluation incorporates all test edges in the $F_1$ calculation to capture the full spectrum of temporal patterns. In contrast, for cause-and-effect and long-range spatio-temporal tasks, we restrict the evaluation to edges involving the memory/target node, as they are the only predictable connections within the otherwise stochastically generated graph. This targeted evaluation ensures that model performance reflects genuine temporal reasoning capabilities rather than chance correlations in random edge formations.


%% file: tex/050periodic.tex
\subsection{Periodicity Tasks}\label{synth-benchmark:periodicity}

We introduce a family of synthetic tasks designed to evaluate temporal graph learning (TGL) methods' ability to recognize periodic structures. These tasks assess two fundamental capabilities: \emph{counting} and \emph{memorization}, in both deterministic and stochastic environments.

\begin{figure}[ht]
    \centering
    \includegraphics[width=\linewidth, height=8em, keepaspectratio]{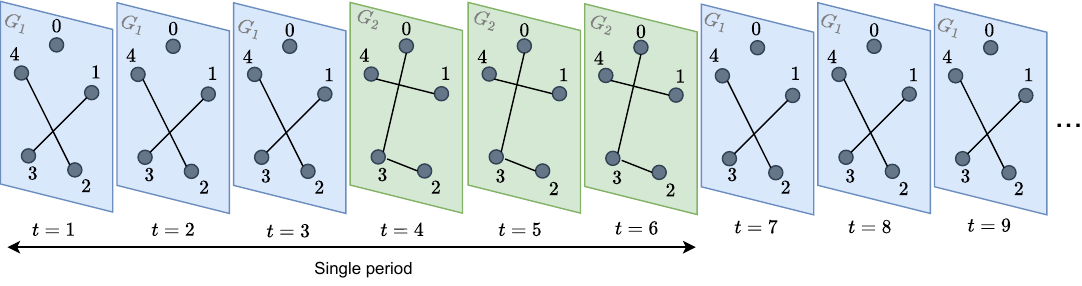}
    \caption{Periodic task in $\Pdet{k\!=\!2,n\!=\!3}$ with 2 unique snapshots repeated 3 times within a  period. }
    \label{fig:periodicity}
\end{figure}

\begin{definition}[Periodicity Tasks]
Let $k, n \in \mathbb{N}$. The task families $\Pdet{k,n}$ and $\Psto{k,n}$ are defined based on a repeating pattern where integers $i=1,\dots,k$ each appears $n$ consecutive times before cycling. For each $t$, let $i_t = (\lfloor t/n\rfloor \mod k) + 1$.
In $\Pdet{k,n}$, the dynamic graph $\mathcal{G} = G_1, G_2, \dots$ is a periodic sequence alternating between $k$ fixed static graphs $G_1, \dots, G_k$, i.e., $G_t = G_{i_t}$.
In $\Psto{k,n}$, each $G_t$ is sampled from one of $k$ distributions $D_1, \dots, D_k$ over static graphs (e.g., Erd\H{o}s-Rényi~(ER)~\cite{erdos1960evolution}, Stochastic Block Model~(SBM)~\cite{holland1983stochastic}), with $G_t \sim D_{i_t}$. The resulting sequence is stochastic, but the distribution pattern follows the  periodic structure of $\Pdet{k,n}$.
\end{definition}

For example, a task in $\Pdet{k\!=\!3,n\!=\!2}$  will correspond to a dynamic graph $ G_1,G_1,G_2,G_2,$ $G_3,G_3,G_1,G_1,$..., where $G_1,G_2,G_3$ are static graphs. A task in $\Pdet{k,n}$ is illustrated in Figure~\ref{fig:periodicity}.

\textbf{Task objectives.}
Tasks in $\Pdet{k,n}$ test the counting and memory capacity of TGL methods. The parameter $k$ controls the length of the pattern and thus the memory demand, while $n$ governs how long each graph is repeated and tests the model's ability to count. For instance, solving a task in $\Pdet{2,n}$ requires counting up to $n$ before switching graphs, while solving a task in $\Pdet{k,1}$ requires memorizing $k$ static graphs. Tasks in $\Psto{k,n}$ introduce stochasticity, requiring models to reason over distributions rather than fixed structures, increasing the complexity while retaining the same periodic structure. Detailed dataset statistics for these tasks can be found in \Cref{appendix:periodicity-deterministic-stats}.

\begin{figure}
    \centering
    \begin{subfigure}[b]{0.3\textwidth}
        \centering
        \includegraphics[width=\linewidth,trim={0 2.25cm 0 0},clip]{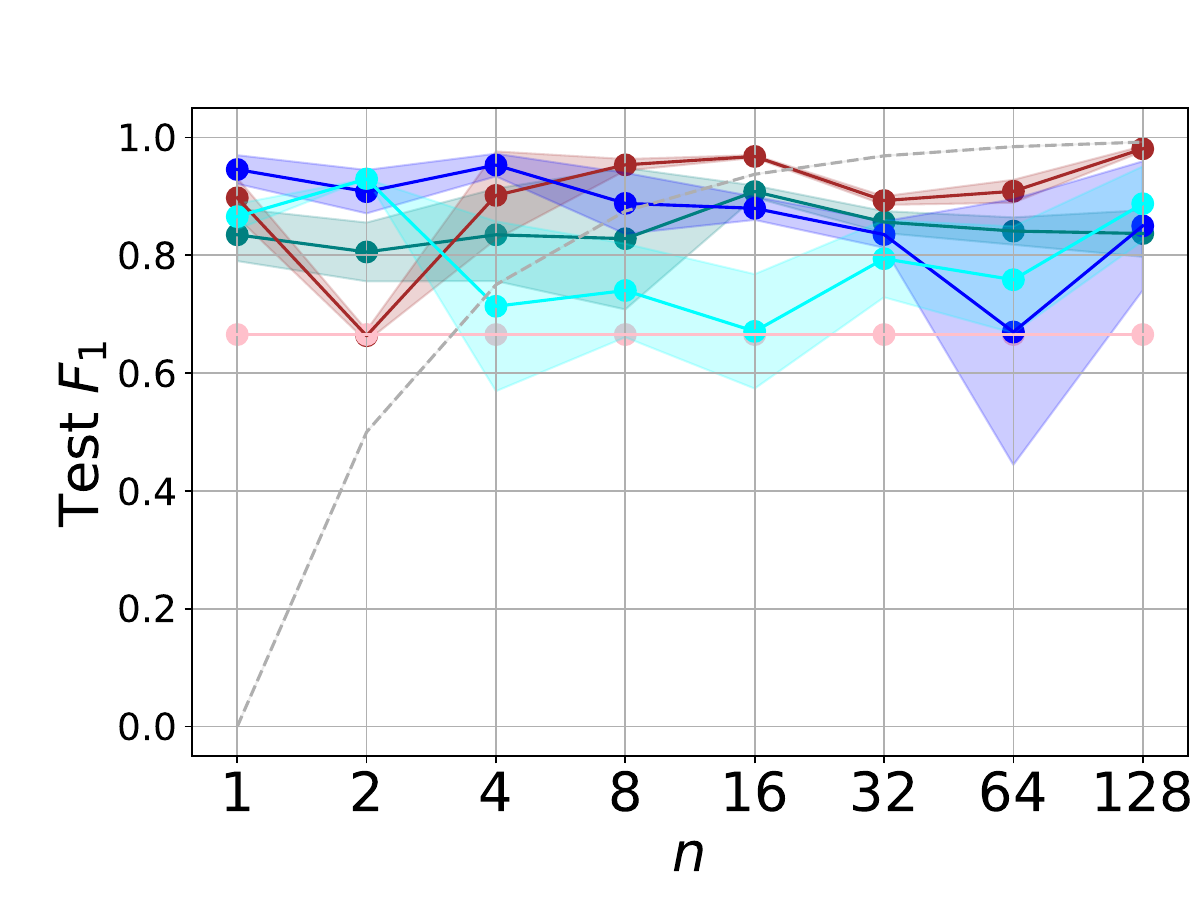}      
        \includegraphics[width=\linewidth,trim={0 0 0 1.2cm },clip]{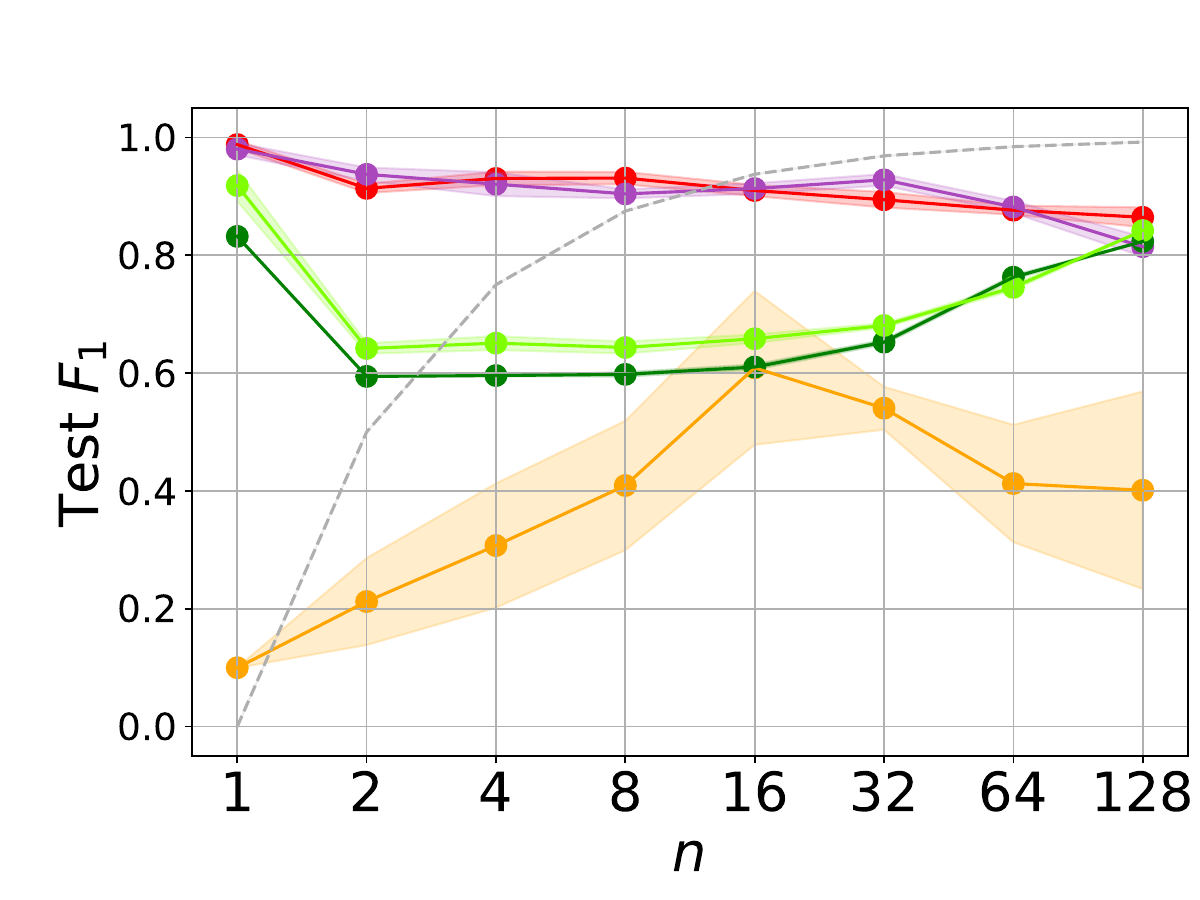}
        \caption{$\Pdet{2,n}$ - All time steps.}
        \label{fig:dtdg_(2,n)_allsteps}
    \end{subfigure}
    \begin{subfigure}[b]{0.3\textwidth}
        \centering
        \includegraphics[width=\linewidth,trim={0 2.25cm 0 0},clip]{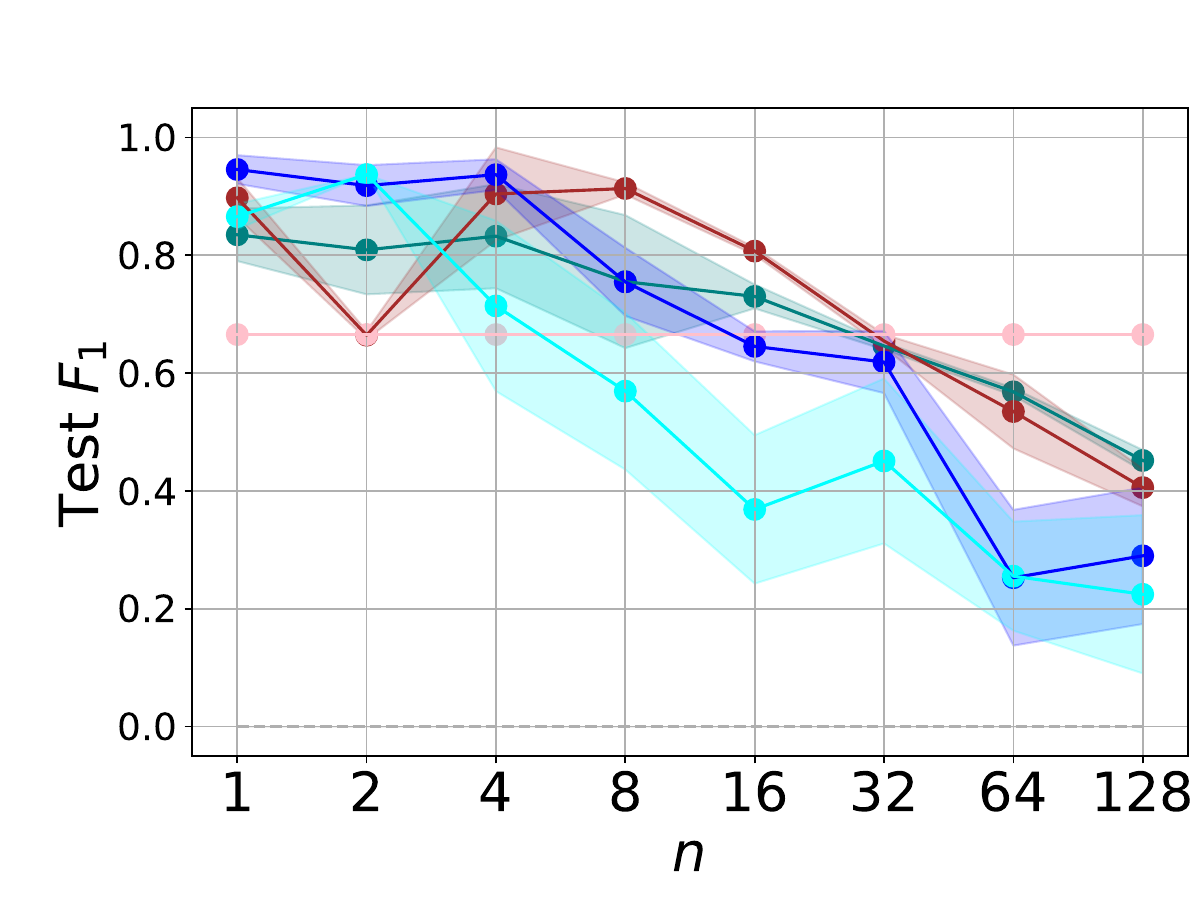}
        \includegraphics[width=\linewidth,trim={0 0 0 1.2cm },clip]{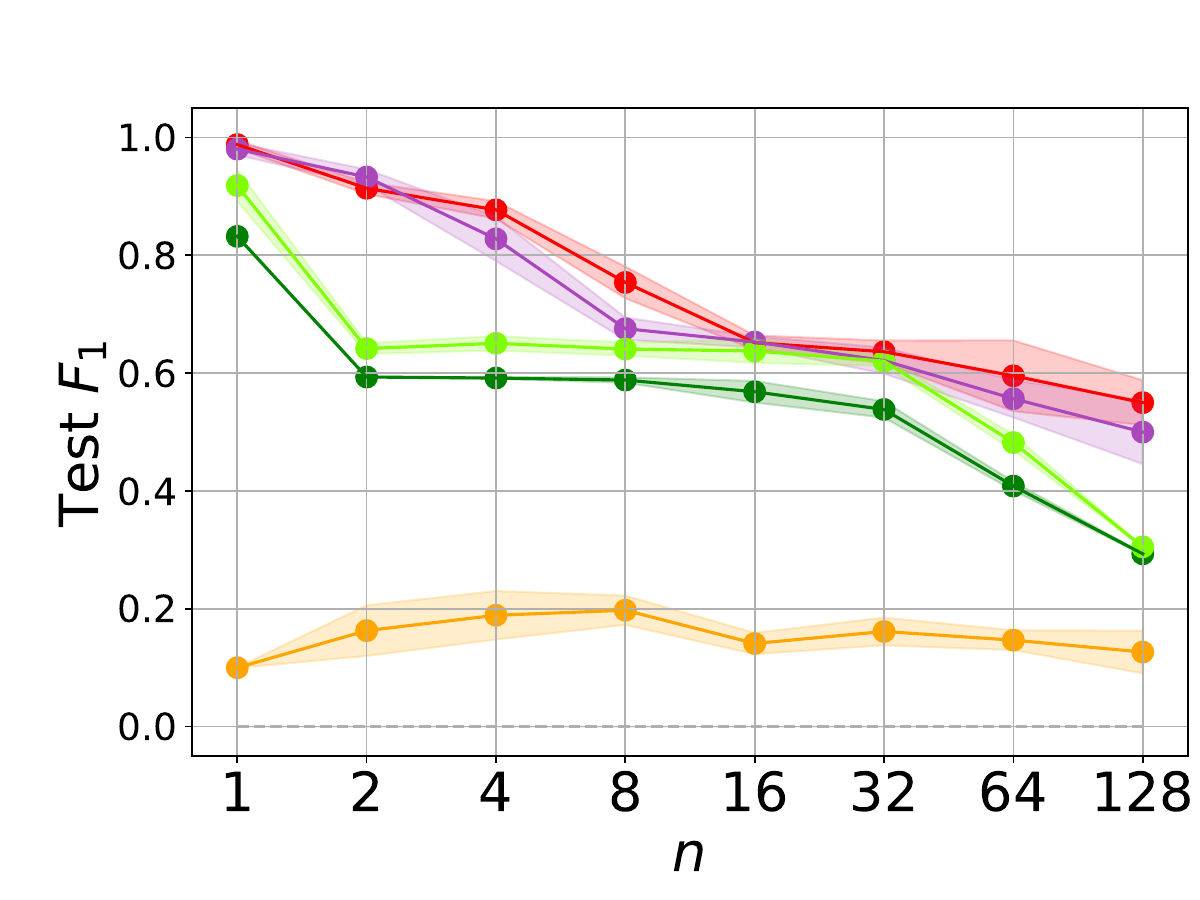}
        \caption{ $\Pdet{2,n}$ - Change points.}
        \label{fig:dtdg_(2,n)_changepoints}
    \end{subfigure}
    \begin{subfigure}[b]{0.3\textwidth}
        \centering
        \includegraphics[width=\linewidth,trim={0 2.25cm 0 0},clip]{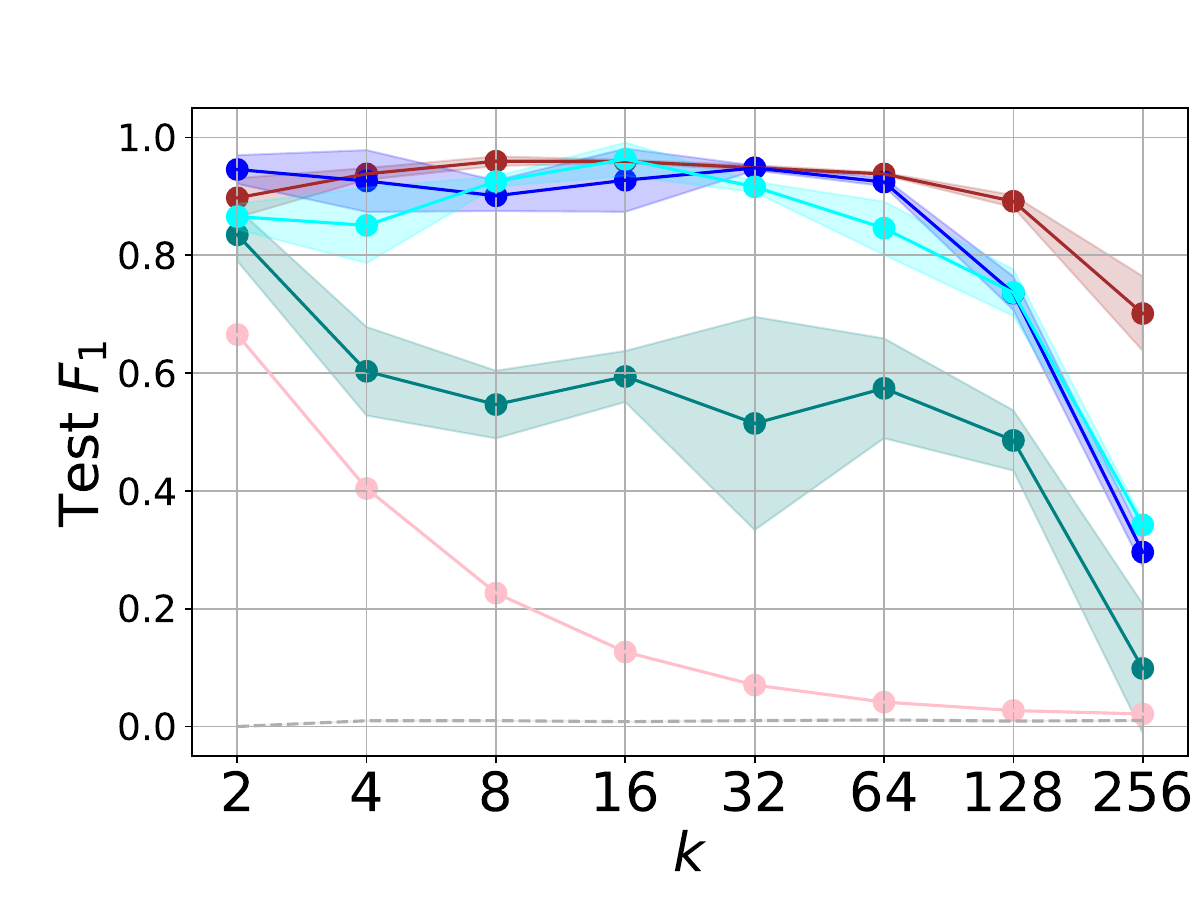}
        
        \includegraphics[width=\linewidth,trim={0 0 0 1.2cm },clip]{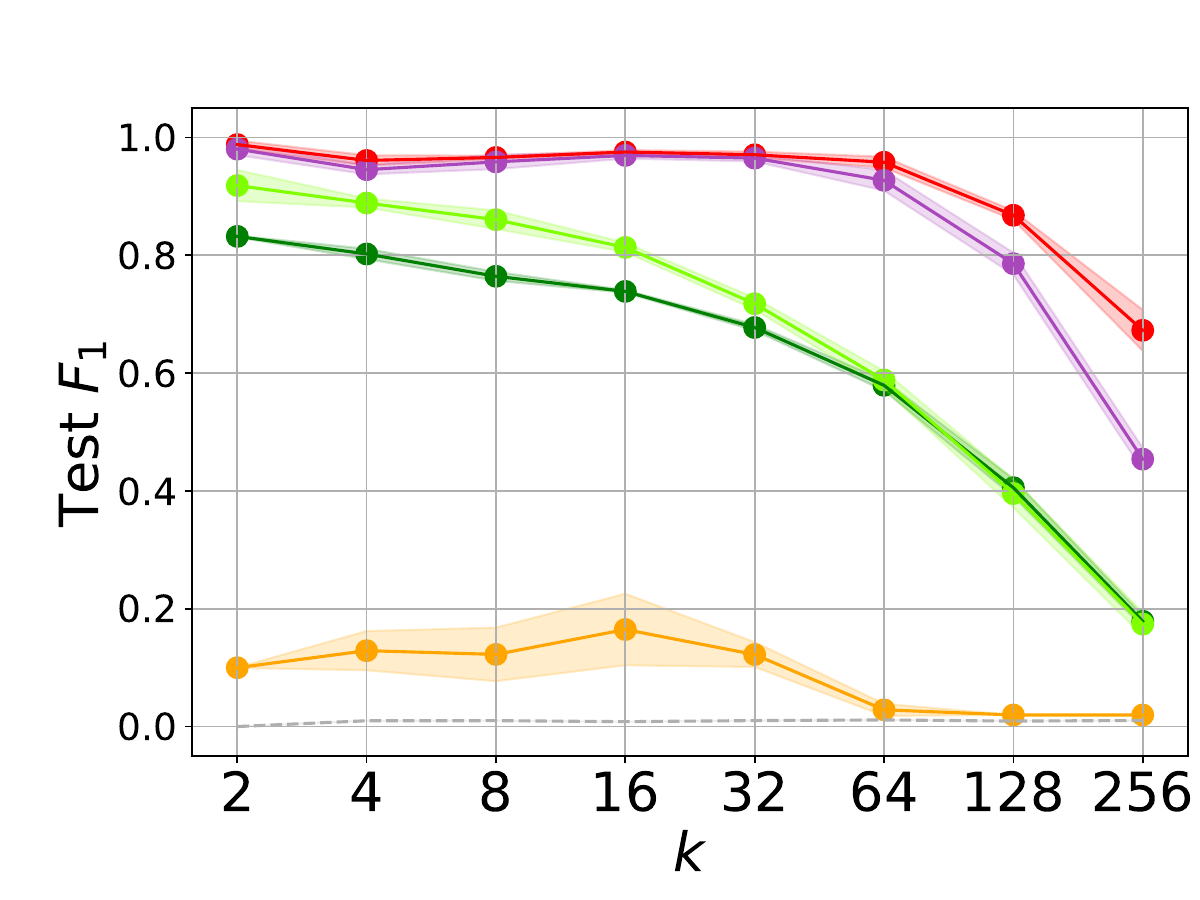}
        \caption{$\Pdet{k, 1}$ - All time steps.}
        \label{fig:dtdg_(k, 1)}
    \end{subfigure}
    \begin{subfigure}[b]{0.08\textwidth}
        \centering
        \raisebox{14mm}{\includegraphics[width=\linewidth]{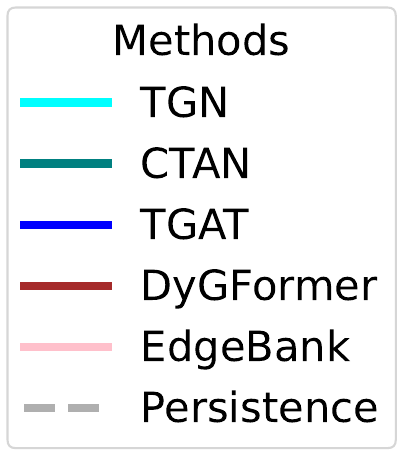}}
        
        \raisebox{14mm}{\includegraphics[width=\linewidth]{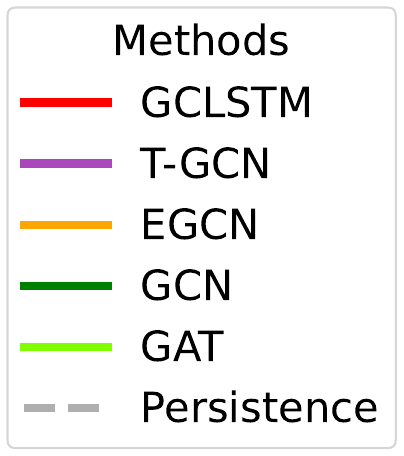}}
    \end{subfigure}
    
    \caption{
    Performance of CTDG methods (top row) and DTDG methods (bottom row) on the deterministic periodicity tasks.
    }
    \label{fig:det_periodic_overall}
\end{figure}

\looseness-1 \textbf{Can TGNNs count?} To evaluate the counting ability of TGNNs, we construct tasks in $\Pdet{k,n}$ with graphs sampled from the Erdős-Rényi (ER) model~\citep{erdos1960evolution} (100 nodes, edge probability 0.01). We set $k = 2$ and vary $n$, where increasing $n$ corresponds to greater task difficulty as models must count longer sequences.

Model performance on these $\Pdet{2,n}$ tasks for $n$ ranging from 1 to 128 is presented in \Cref{fig:det_periodic_overall}. 
Since models that simply repeat the previous timestep can perform well for large $n$, we evaluate both overall performance~(left plot) and performance at change points~(center plot) where the active graph switches (i.e., $t = n+1, 2n+1, \cdots$). High scores at change points indicate true counting and pattern understanding, whereas good overall performance with poor change point performance suggests the model is merely exploiting continuity rather than reasoning about periodicity.


\looseness-1 The results reveal distinct behaviors across different TGNN families. Among CTDG models, TGAT excels for small $n$, while DyGFormer demonstrates more consistent performance at larger $n$. For DTDG methods, T-GCN and GC-LSTM are strongest overall. Notably, EvolveGCN performs worse than static graph learning methods (GCN and GAT), which lack temporal processing mechanisms, underscoring how challenging these seemingly simple periodic patterns can be for current TGNNs. EdgeBank remains constant across all $n$ values, always predicting the union of both graphs. The persistence baseline improves over all timesteps as $n$ increases by simply copying the previous snapshot, but scores zero at change points.

\looseness-1 A key observation, evident when comparing performance over all timesteps versus at change points (\Cref{fig:det_periodic_overall}), is that as $n$ grows, many models increasingly rely on repetition rather than explicit counting. While their overall scores might remain high or even improve for larger $n$ (due to successfully predicting links during long static phases), their performance at change points often degrades. At $n = 32$, EdgeBank even starts to outperform all TGNNs. This divergence suggests that current TGNNs struggle to robustly count long sequences and instead learn a simpler heuristic reminiscent of persistence. Further per-timestep analysis supporting this trend is provided in Appendix~\ref{appendix:pertimesteps_counting}.

\looseness-1 \textbf{How much can TGNNs memorize?} To evaluate the memorization capabilities of TGNNs, we fix $n = 1$ and vary $k$. The difficulty scales with $k$: as more unique graph structures are introduced, models must maintain a larger and more distinct set of representations to correctly predict links at each timestep. \label{seq:memorize}

\looseness-1 Figure~\ref{fig:det_periodic_overall} (right) shows results on $\Pdet{k, 1}$ for $k$ ranging from 2 to 256. As expected, EdgeBank steadily degrades to zero performance at $k = 256$, eventually defaulting to predicting each snapshot as a clique. All TGNN models show a gradual decline as $k$ increases. GC-LSTM and DyGFormer consistently perform best, demonstrating strong memorization of patterns. T-GCN, TGN, and TGAT remain robust up to $k = 128$, but drop sharply at $k = 256$, suggesting these models reach their maximum memorization capacity at this point.

\looseness-1 In contrast, TGNNs such as CTAN and EvolveGCN struggle significantly as the number of unique graphs increases. CTAN's performance begins to degrade considerably after $k=64$. Notably, both these methods are often outperformed by static GNNs like GCN and GAT, which exhibit a more gradual decline. This suggests that for tasks dominated by the need to memorize distinct states, an ineffective temporal mechanism can be more detrimental than no temporal mechanism at all.

\looseness-1 \textbf{Can TGNNs learn stochastic periodic structures?} Finally, we investigate how models' memorization capabilities extend to probabilistic settings, where periodic structure emerges from stochastic processes rather than deterministic patterns. 
We employ Stochastic Block Models (SBMs) with 100 nodes divided into 3 communities. While all SBM distributions share identical inter-community (0.01) edge probabilities, they differ in community structures (details in Appendix~\ref{appendix:periodicity_stochastic_design}). We examine two intra-community edge probability settings: $p=0.9$ and $p=0.5$~(the latter being  more difficult). 

Figure~\ref{fig:(K,1)_overall_sto} presents model performance across these stochastic periodicity tasks. As expected, performance decreases as $k$ increases, reflecting the growing challenge of memorizing multiple stochastic patterns. GC-LSTM consistently achieves the highest performance across both probability settings, with TGAT as the second-best performer. While T-GCN performs strongly with clearer community structure ($p=0.9$), it struggles in the noisier setting ($p=0.5$), falling behind TGAT and TGN. Overall, performance decreases under lower intra-community density, confirming that increased stochasticity challenges memorization capabilities. Notably, unlike in deterministic scenarios, static baselines (GCN and GAT) consistently underperform temporal models, highlighting the effectiveness of temporal modeling in capturing stochastic periodic structures within dynamic graphs.

\begin{figure}
    \centering
        \includegraphics[width=0.35\linewidth]{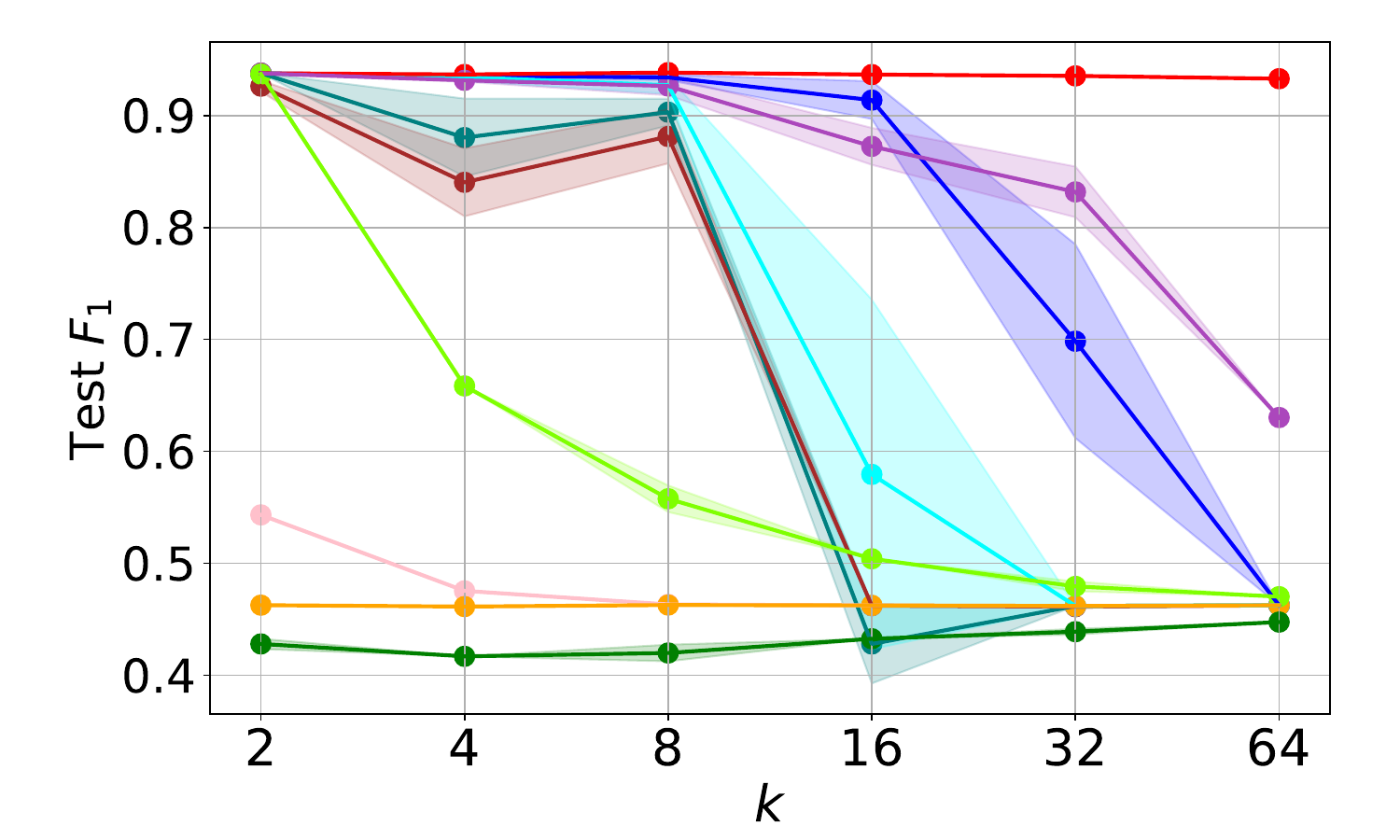}%
        \hspace{0.5cm}%
        \includegraphics[width=0.35\linewidth]{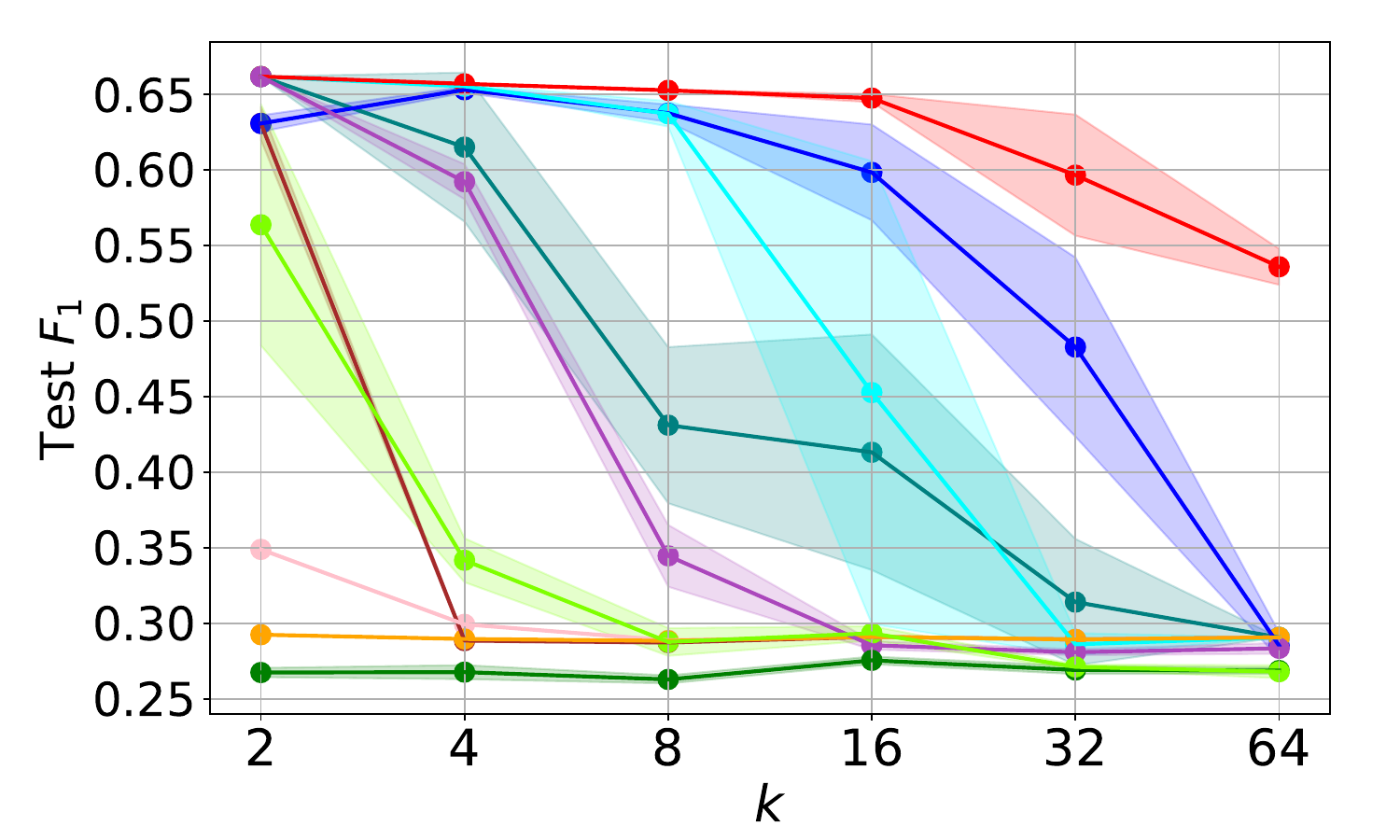}%
        \raisebox{5mm}{
        \includegraphics[width=0.08\linewidth]{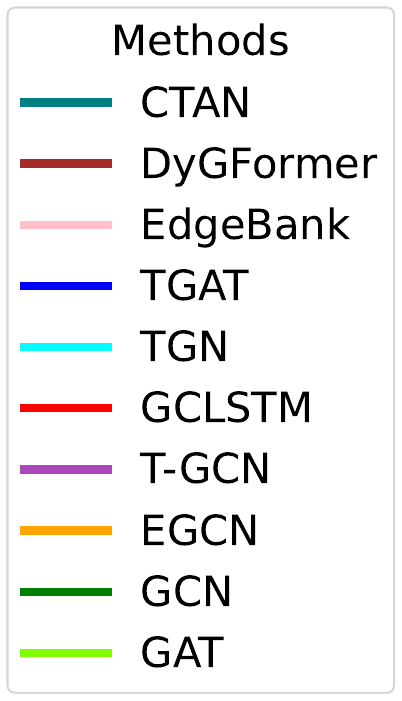}}
    \caption{ Performance of TGL methods on the stochastic periodicity tasks~$\Psto{k,1}$ for different values of the intra-community edge probability: $p=0.9$ (left) and $p=0.5$ (right).
    }
    \label{fig:(K,1)_overall_sto}
\end{figure}

%% file: tex/060cause.tex
\subsection{Delayed Cause-and-Effect Tasks} \label{synth-benchmark:cause}

\looseness-1 To assess how effectively TGL methods capture delayed causal relationships, we introduce \emph{delayed cause-and-effect tasks}. These tasks involve a sequence of randomly generated graphs (e.g., from an Erd\H{o}s-Rényi distribution) with a designated memory node that connects to nodes participating in edges from \(\lag\) time steps in the past. This design creates a clear temporal dependency that models must identify. We formalize this framework as follows:

\begin{definition}[Delayed Cause-and-Effect Task]
Let \(\lag \in \mathbb{N}\). The \emph{delayed cause-and-effect task} family \(\causeeffect{\lag}\) consists of dynamic graphs \(\mathcal{G} = G_1, G_2, \dots\) generated as follows: First, each graph \(G_t = (V, E_t)\) is sampled independently from a distribution \(D\) over static graphs, where \(V = \{v_1, \dots, v_N\}\) is the set of nodes, shared across time steps. For \(t > \lag\), the graph \(G_t\) is augmented by introducing a memory node \(v_{\mathcal M}\) and connecting it to the nodes involved in edges at time \(t - \lag\):
\[
V \leftarrow V \cup \{v_{\mathcal M}\}, \quad E_t \leftarrow E_t \cup \left\{ (v_{\mathcal M}, u), (v_{\mathcal M}, v) \mid (u,v) \in E_{t-\lag} \right\}  \text{ for } t > \lag.
\]
\end{definition}

 \begin{figure}[t]
    \centering
    \includegraphics[width=0.8\linewidth, height=8em, keepaspectratio]{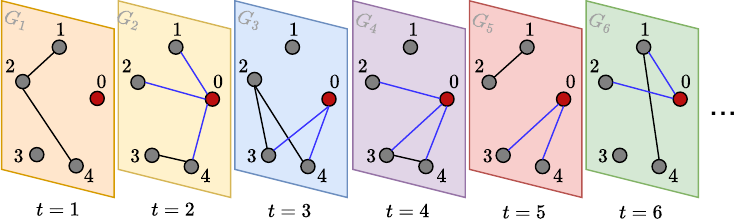}
    \caption{Illustration of a delayed cause-and-effect dataset with lag $\lag=1$. \underline{Black edges} show the cause subgraph (nodes 1 to $N$). \underline{\textcolor{blue}{Blue edges}} show the effect subgraph, where the memory node (0) connects to previously active cause nodes (degree $\geq$ 1). }
    \label{fig:memnode}
\end{figure}

\looseness-1 Intuitively, the model needs to remember nodes that were connected to each other at $E_{t-\ell}$ to predict the edges of the memory node at time $E_t$. An example of this task is illustrated in Figure~\ref{fig:memnode}. As \(\lag\) increases, the task becomes more challenging in terms of the memory capacity required from the model. While the number of nodes and edges in the graphs \(G_t\) also affect the task's difficulty, we focus on varying \(\lag\). Since only the edges involving the memory node can be predicted~(the other ones being completely random), only the $F_1$ over possible edges involving the memory node are reported in our experiments.

\looseness-1 \textbf{Task objectives and implementation.} In the delayed cause-and-effect tasks \(\causeeffect{\lag}\), TGNNs must propagate information across time steps while identifying the causal relationship governing the memory node's connectivity. This requires models to recognize temporal patterns and maintain historical information. The task difficulty scales with \(\lag\): larger values require retaining information longer, making causal relationship identification increasingly challenging. For implementation, we generate the underlying graphs (excluding the memory node) using an Erdős-Rényi model with 100 nodes and edge probability 0.01, creating a controlled environment to isolate and evaluate temporal reasoning capabilities.

\begin{wrapfigure}{l}{0.5\textwidth}
    \includegraphics[width=0.7\linewidth]{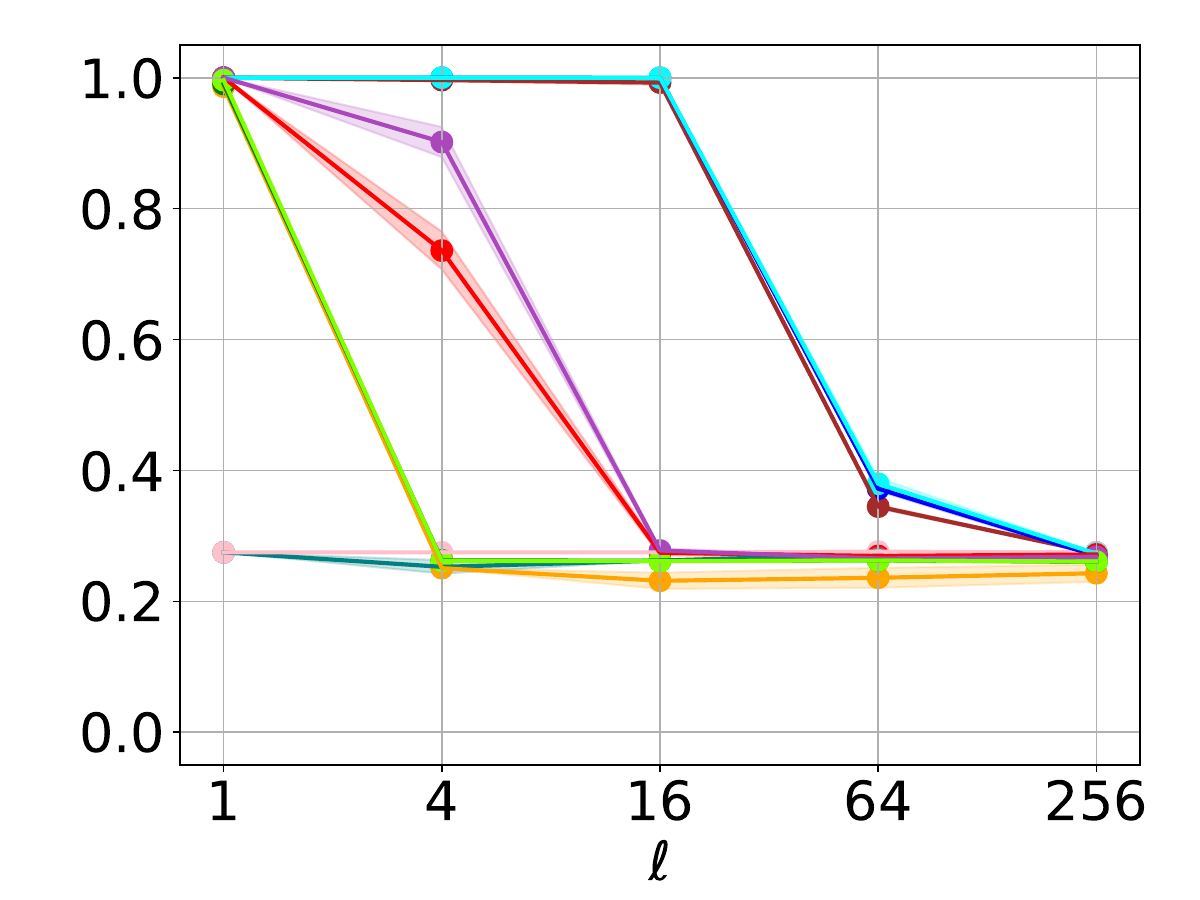}
    \raisebox{5mm}{\includegraphics[width=0.2\linewidth]{figs/legends/CTDG_DTDG_all_w_edgebank_wo_persistence.pdf}}
        \caption{Methods' performance on cause-and-effect tasks $\causeeffect{\ell}$ across five temporal lags $\ell$.}
    \label{fig:plot_mem_node}
\end{wrapfigure}

\looseness-1 \textbf{Results.} Figure~\ref{fig:plot_mem_node} illustrates performance on cause-and-effect tasks across varying temporal lags. For minimal lags ($\ell = 1$), most methods demonstrate strong performance, as the task primarily requires extracting information from the immediate past, a capability even static models possess. However, as lag increases, performance degrades systematically, with static and DTDG approaches struggling significantly with longer temporal dependencies. GC-LSTM and T-GCN exhibit particularly pronounced performance deterioration at $\ell = 4$ and $\ell = 16$, revealing limitations in recurrent architectures' capacity to maintain long-term temporal information. Conversely, DyGFormer, TGAT, and TGN consistently outperform other methods, underscoring the efficacy of attention mechanisms and memory modules in capturing dynamic temporal patterns. At extended lags ($\ell = 64$ and $\ell = 256$), all methods converge toward EdgeBank's performance level, indicating a common challenge in modeling distant causal relationships. Notably, CTAN shows inconsistent performance even at minimal lags, suggesting inherent limitations in this context. These findings emphasize the crucial importance of architectural design choices, particularly attention and memory components, for effective temporal reasoning in dynamic graphs.

%% file: tex/070long.tex
\subsection{Long-Range Spatio-Temporal Task} \label{synth-benchmark:long}

\looseness-1  Finally, we introduce \emph{long-range spatio-temporal tasks} to evaluate how effectively TGL methods reason across both temporal and spatial dimensions. These tasks extend the delayed cause-and-effect framework by incorporating multi-hop spatial paths alongside temporal dependencies. In this setting, each graph snapshot contains multiple paths originating from a source node \(v_{\mathcal S}\), while a target node \(v_{\mathcal T}\) connects to the endpoints of paths appearing \(\lag\) time steps earlier. We formalize this task as follows:
 
\begin{definition}[Long-Range Spatio-Temporal Task]
\looseness-1 Let \(\lag, d, P \in \mathbb{N}\). The \emph{long-range spatio-temporal task} family $ \mathcal{L}\hspace{-0.035cm}\mathcal{R}_P(\lag,d)$ is defined over dynamic graphs \(\mathcal{G} = G_1, G_2, \dots\), where each snapshot $G_t = (V, E_t)$ has a fixed node set $V = \{v_{\mathcal S}, v_{\mathcal T}, v_1, \dots, v_{N}\}$. The edge set $E_t$ consists of $P$ disjoint paths of length $d$ from the source node $v_{\mathcal S}$ through randomly chosen intermediate nodes. For $t > \lag$, $E_t$ additionally includes $P$ edges from the target node $v_{\mathcal T}$ to the endpoints of the $P$ paths in $G_{t - \lag}$. Formally, for $t > \lag$:
$
E_t = \bigcup_{p=1}^P \{(v_{\mathcal S}, u_1^{(t,p)}), (u_1^{(t,p)}, u_2^{(t,p)}),\dots, (u_{d-1}^{(t,p)}, u_d^{(t,p)})\} \cup \{(v_{\mathcal T}, u_d^{(t-\lag,p)})\}$, where the nodes $u^{(t,p)}_{i}$ for $1\leq i\leq d$, $1\leq p \leq P$ are drawn at random in $\{v_1,\dots,v_N\}$~(without replacement).
\end{definition}\label{def:long_range}

\begin{figure}
    \centering
    \includegraphics[width=0.8\linewidth, height=11.0em, keepaspectratio]{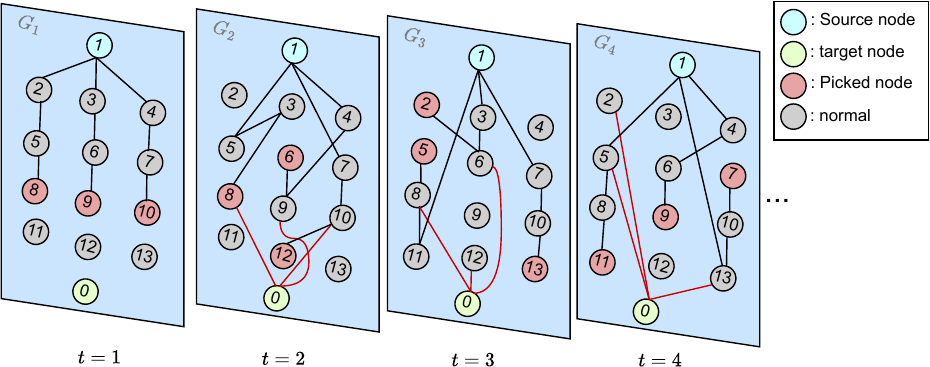}
    \caption{Illustration of a long-range spatio-temporal task $\longrange{\lag,d}$ with    $\ell=1$, $d=3$ and $P=3$.}
    \label{fig:longrange}
\end{figure}

Intuitively, we can treat node $v_\mathcal{S}$ as the \textit{progenitor} of some signal that 1) reaches a set of nodes that are spatially separated from $v_\mathcal{S}$ by a distance $d$ and 2) whose effect (connecting the nodes that it reached to node $v_\mathcal{T})$ is additionally delayed by $\lag$ timesteps.
An illustration of a dynamic graph for this task is given in Figure~\ref{fig:longrange}.
We will focus on the effect of the lag $\lag$  and distance $d$ in our experiments and set the number of path $P$ to $3$ in all tasks, which we will simply denote by $\longrange{\lag,d}$.

\textbf{Task objectives and implementation.}
Dynamic graphs inherently contain two fundamental distance metrics: temporal (between time steps) and spatial (between nodes). Effective TGL methods must reason across both dimensions simultaneously to capture complex patterns. The \(\longrange{\lag,d}\) task family specifically evaluates this capability by requiring models to track information across both temporal lags and multi-hop spatial paths. The parameter \(\lag\) determines the temporal memory requirement, while \(d\) defines the spatial propagation distance. This dual-parameter design creates a comprehensive benchmark for assessing spatio-temporal reasoning depth in TGL architectures. To systematically evaluate model capabilities, we generate tasks with increasing complexity using distance values $d \in \{1, 2, 4, 8, 16\}$ and temporal lags $\lag \in \{1, 4, 16, 32\}$.

\begin{figure}[t]
    \centering
    \begin{subfigure}[b]{0.24\textwidth}
        \centering
        \includegraphics[width=\linewidth]{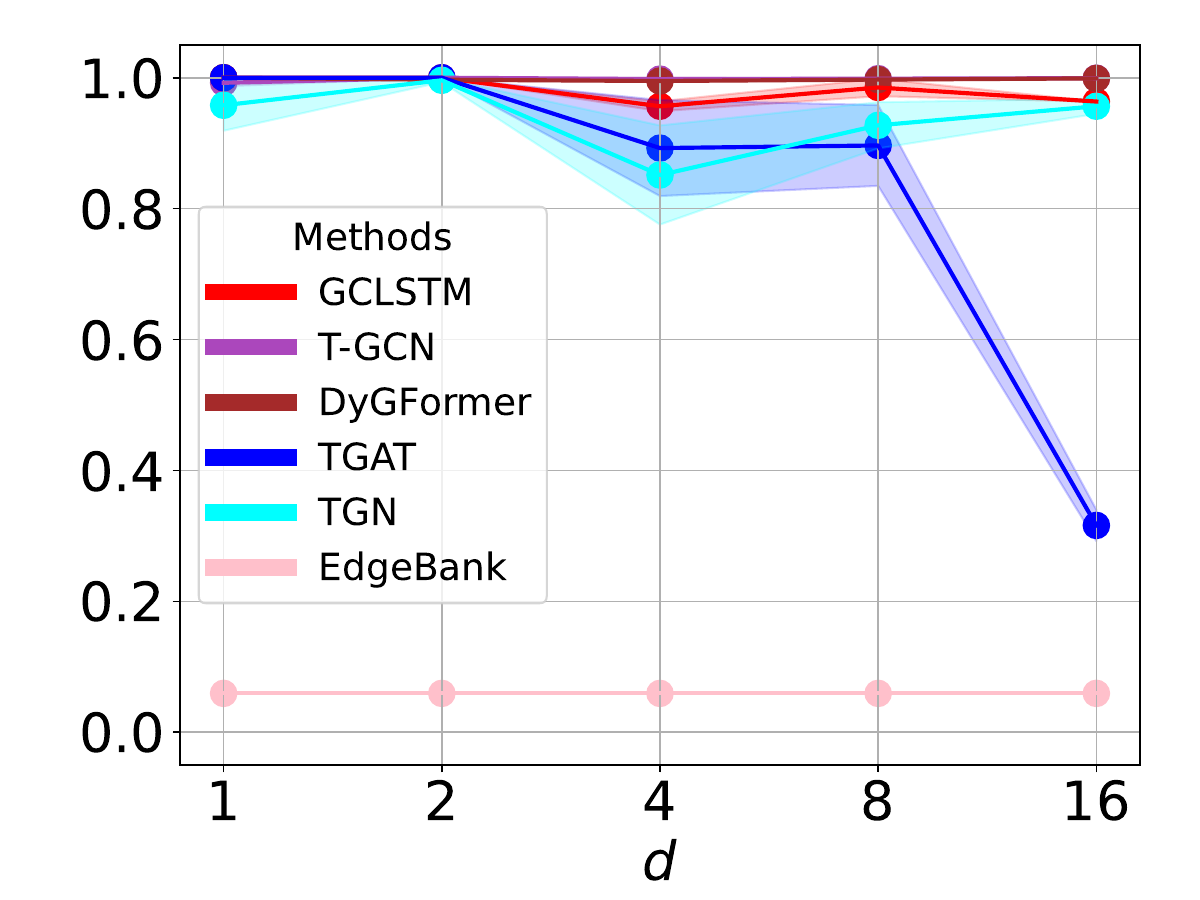}
        \caption{$(\ell=1, d)$}
        \label{fig:sub_long_range_(1,d)}
    \end{subfigure}
    \begin{subfigure}[b]{0.24\textwidth}
        \centering
        \includegraphics[width=\linewidth]{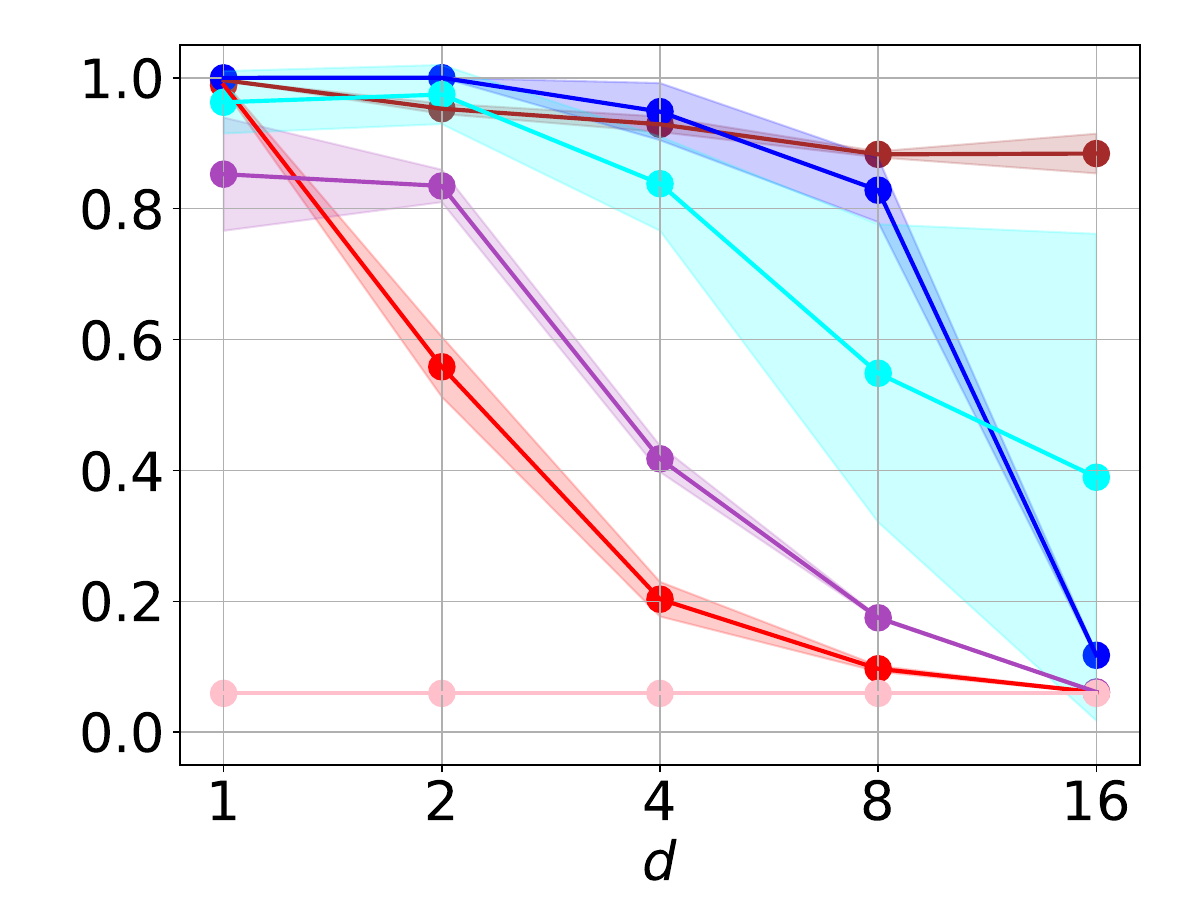}
        \caption{$(\ell=4, d)$}
        \label{fig:sub_long_range_(4,d)}
    \end{subfigure}
    \centering
    \begin{subfigure}[b]{0.24\textwidth}
        \centering
        \includegraphics[width=\linewidth]{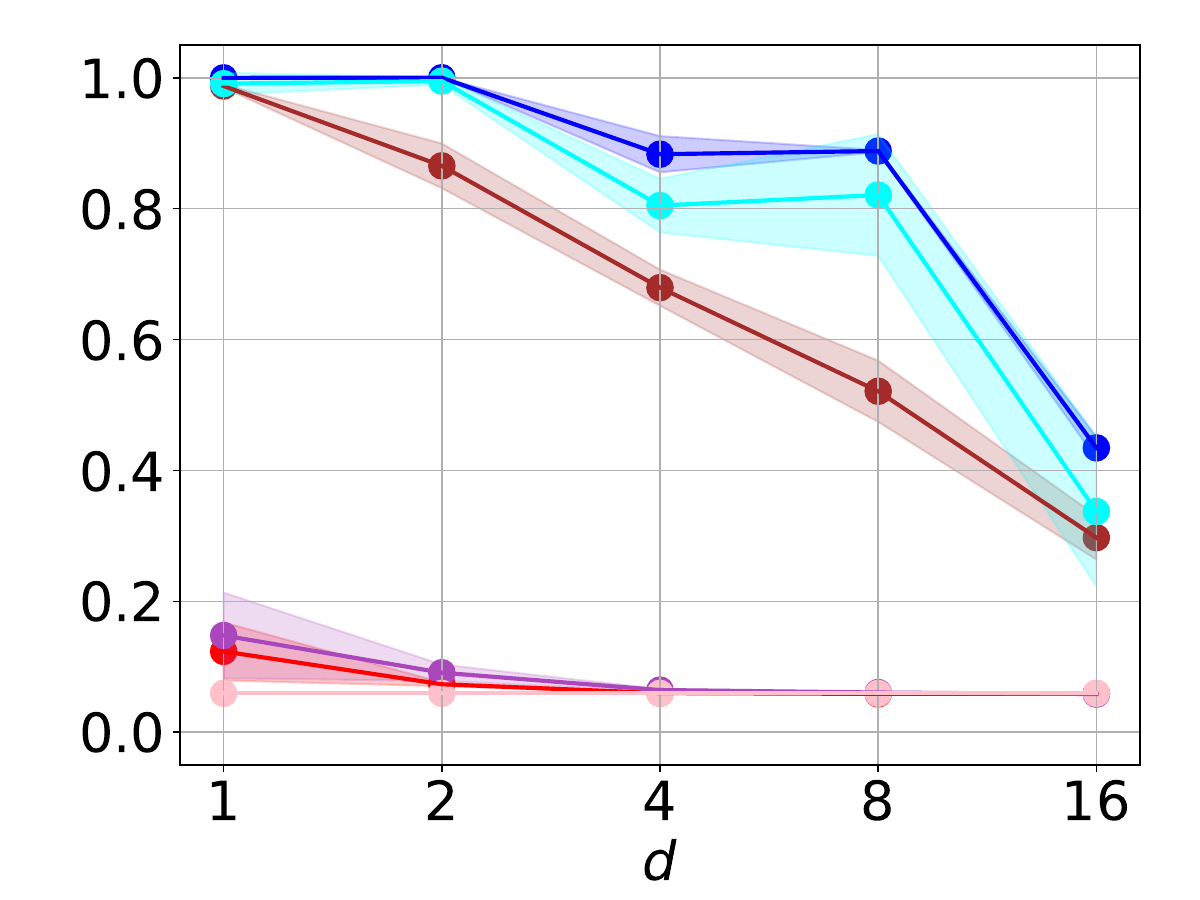}
        \caption{$(\ell=16, d)$}
        \label{fig:sub_long_range_(16,d)}
    \end{subfigure}
    \begin{subfigure}[b]{0.24\textwidth}
        \centering
        \includegraphics[width=\linewidth]{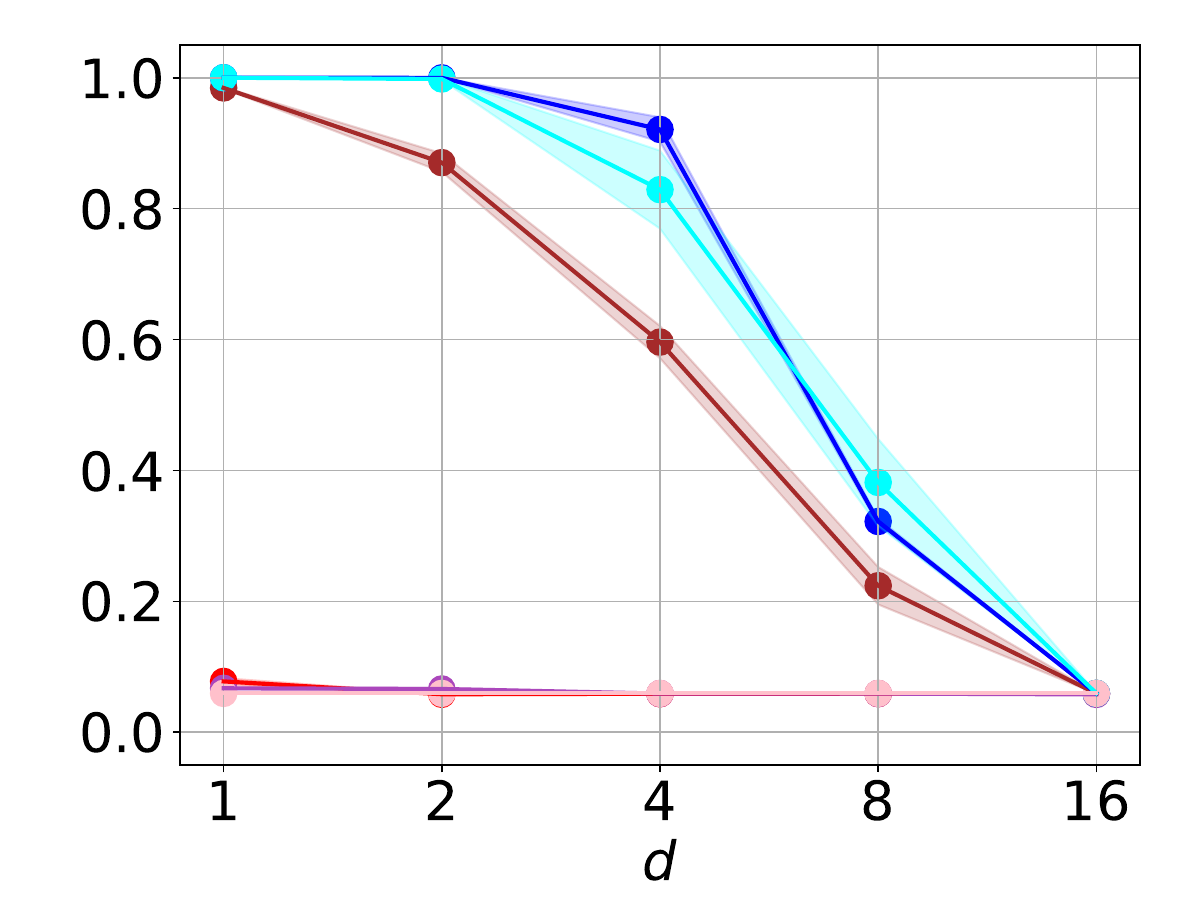}
        \caption{$(\ell=32, d)$}
        \label{fig:sub_long_range_(32,d)}
    \end{subfigure}
    \caption{Performance varying spatial and temporal distances in the long-range spatio-temporal tasks.}
    \label{fig:plots_long_range}
\end{figure}

\textbf{Results.} Figure~\ref{fig:plots_long_range} presents $F_1$ scores for top-performing methods and the EdgeBank baseline (complete results in Appendix~\ref{apendix:spatio_temporal_full_methods}). At small temporal lags, both DTDG approaches and DyGFormer demonstrate strong performance. However, as $\lag$ increases, DTDG methods exhibit sharp performance declines, revealing limitations in recurrent architectures' capacity to maintain long-range temporal memory. DyGFormer's performance deteriorates more gradually but still struggles with extended temporal dependencies. In contrast, TGAT and TGN maintain robust performance even at substantial lags ($\lag=16,32$), highlighting the efficacy of attention-based message passing for complex spatio-temporal reasoning. Nevertheless, all models, even the strongest temporal reasoners, show significant performance degradation when spatial distance exceeds $d=8$, emphasizing the persistent challenge of capturing long-range spatial dependencies in dynamic graphs.

%% file: tex/080discussion.tex
\section{Discussion and Analysis}\label{sec:discussion}

\begin{table}[t]
\centering
\caption{Average method rank~($\downarrow$) for \benchmarkname tasks~(c.p. stands for change points). Top results are shown in \first{first}, \second{second}, \third{third}.} \label{tab:overall}
\resizebox{\linewidth}{!}{%
\begin{tabular}{l l |ccccc|c|cccc}
\toprule

&& \multicolumn{5}{c|}{Periodicity} & Cause \& & \multicolumn{4}{c}{Spatio-temporal long-range} \\
& Method & \multicolumn{2}{c}{$\Pdet{2,n}$} &\multirow{2}{*}{$\Pdet{k,1}$} & \multicolumn{2}{c|}{$\Psto{2,n}$} & Effect &  \multicolumn{4}{|c}{$\longrange{\lag,d}$} \\    
& & all $t$  & c.p. &  & $p=0.5$ & $p=0.9$ & $\causeeffect{\ell}$ & $\lag\!=\!1$ & $\lag\!=\!4$ & $\lag\!=\!16$ & $\lag\!=\!32$ \\ 
%
%
  \midrule

  \multirow{4}{*}{\rotatebox[origin=c]{90}{CTDG}} & 
  CTAN & 5.625 & \third{3.75} & 7.625 & \second{3.167} & 5.167 & 6.4 & 10.0 & 8.6 & 8.4 & 6.8 \\
  & DyGFormer & \second{3.25} & \second{3.5} & \third{2.625} & 7.167 & 7.5 & \third{3.2} & \second{2.2} & \second{1.8} & \third{3.0} & \second{3.4} \\
  & TGAT & 4.75 & 5.0 & 4.375 & \third{3.667} & \second{3.167} & \second{2.6} & 3.8 & \first{1.6} & \first{1.0} & \first{1.8} \\
  & TGN & 5.75 & 7.125 & 4.625 & \third{3.667} & 4.33 & \first{2.4} & 4.8 & \third{2.8} & \second{2.0} & \first{1.8} \\
  \midrule
  \multirow{5}{*}{\rotatebox[origin=c]{90}{DTDG}} 
  & GC-LSTM & \first{2.875} & \first{3.125} & \first{1.25} & \first{1.0} & \first{1.167} & 4.8 & \third{3.0} & 4.6 & 5.6 & 6.2 \\
  & EGCN & 10.875 & 10.0 & 9.75 & 6.5 & 8.167 & 5.0 & 8.2 & 7.8 & 7.0 & 8.0 \\
  & T-GCN & \second{3.25} & \third{3.75} & \second{2.5} & 6.5 & \third{3.333} & 9.6 & \first{1.8} & 4.2 & 4.8 & \third{6.0}\\
  & GCN & 8.875 & 8.0 & 7.0 & 9.833 & 9.833 & 7.8 & 6.8 & 8.8 & 9.2 & 9.0 \\
  & GAT & 7.625 & 6.5 & 6.0 & 7.5 & 5.333 & 8.0 & 5.6 & 8.6 & 8.2 & 6.4 \\
  \midrule
  & EdgeBank & 8.5 & 4.25 & 9.25 & 6.0 & 7.0 & 5.2 & 8.8 & 6.0 & 5.4 & 9.8 \\
  & Persistence & \third{4.625} & 11.0 & 11.0 & 11.0 & 11.0 & 11.0 & 11.0 & 11.0 & 11.0 & 11.0 \\
  \bottomrule
\end{tabular}
}
\end{table}

In this section, we synthesize key findings and insights across all three task categories in \benchmarkname.

\textbf{Bird's eye view of the results.} Table~\ref{tab:overall} presents a comprehensive ranking of methods across our benchmark tasks. 
Our analysis reveals a striking dichotomy in model effectiveness across different temporal reasoning challenges. For periodicity tasks, GC-LSTM consistently outperforms all competitors across all five settings, challenging the prevailing assumption that CTDG methods outperform DTDG ones~\cite{huangutg}. This suggests that for tasks requiring precise counting and pattern memorization, the simpler recurrent architecture of DTDG models may offer computational advantages over their more complex continuous-time counterparts. Conversely, for cause-and-effect and spatio-temporal long-range tasks, CTDG methods, particularly TGN and TGAT emerge as the dominant methods. Notably, CTAN, despite being explicitly designed for long-range temporal propagation, performs surprisingly poorly on spatio-temporal long-range tasks. This unexpected result highlights potential limitations in its ability to \textit{jointly reason} over spatial and temporal dimensions.

\looseness-1 These findings collectively demonstrate that no single architectural paradigm excels across all \benchmarkname tasks. The observed performance variations suggest that different architectural inductive biases are better suited to specific temporal reasoning challenges. This insight points to a promising research direction: developing hybrid architectures that can effectively combine the strengths of different approaches to achieve superior performance across diverse temporal reasoning scenarios.

\looseness-1 \textbf{Effect of Number of Neighbors.} Temporal neighbor sampling is a fundamental mechanism in CTDG methods~\cite{dygformer,tgn,tgat,luo2022neighborhood}, that allows  models to access historical interactions and model temporal dependencies, especially crucial in scenarios with sparse edge distributions. Despite its importance, the number of sampled neighbors remains an under-examined hyperparameter, commonly fixed at default values (e.g., 20 neighbors in established literature~\cite{dygformer}). We maintained this conventional setting in our previous experiments but now systematically investigate how this parameter influences performance across \benchmarkname tasks.

\begin{figure}[t]
    \centering
    \begin{subfigure}[b]{0.24\textwidth}
        \centering
        \includegraphics[width=\linewidth]{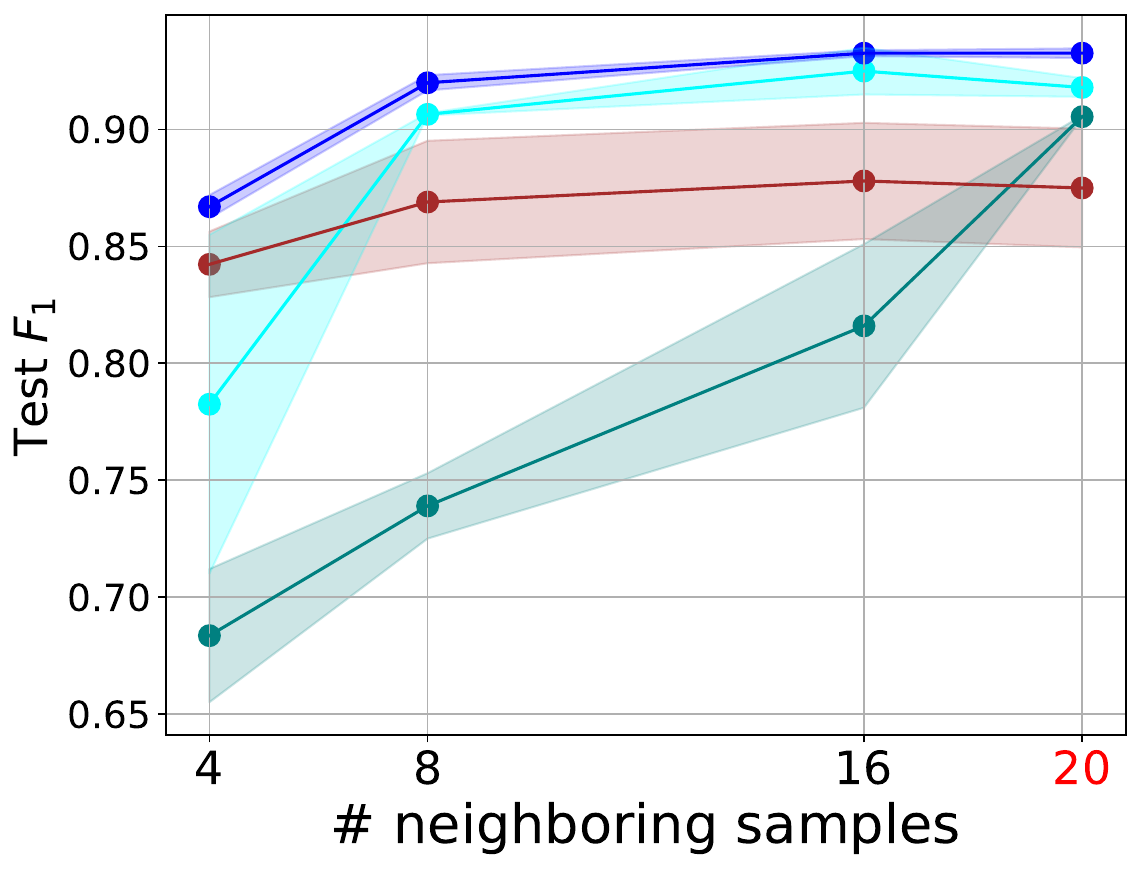}
        \caption{$\Psto{8, 1}$}
        \label{fig:sto_periodic_8_num_nei}
    \end{subfigure}
    \hfill
    \begin{subfigure}[b]{0.232\textwidth}
        \centering
        \includegraphics[width=\linewidth]{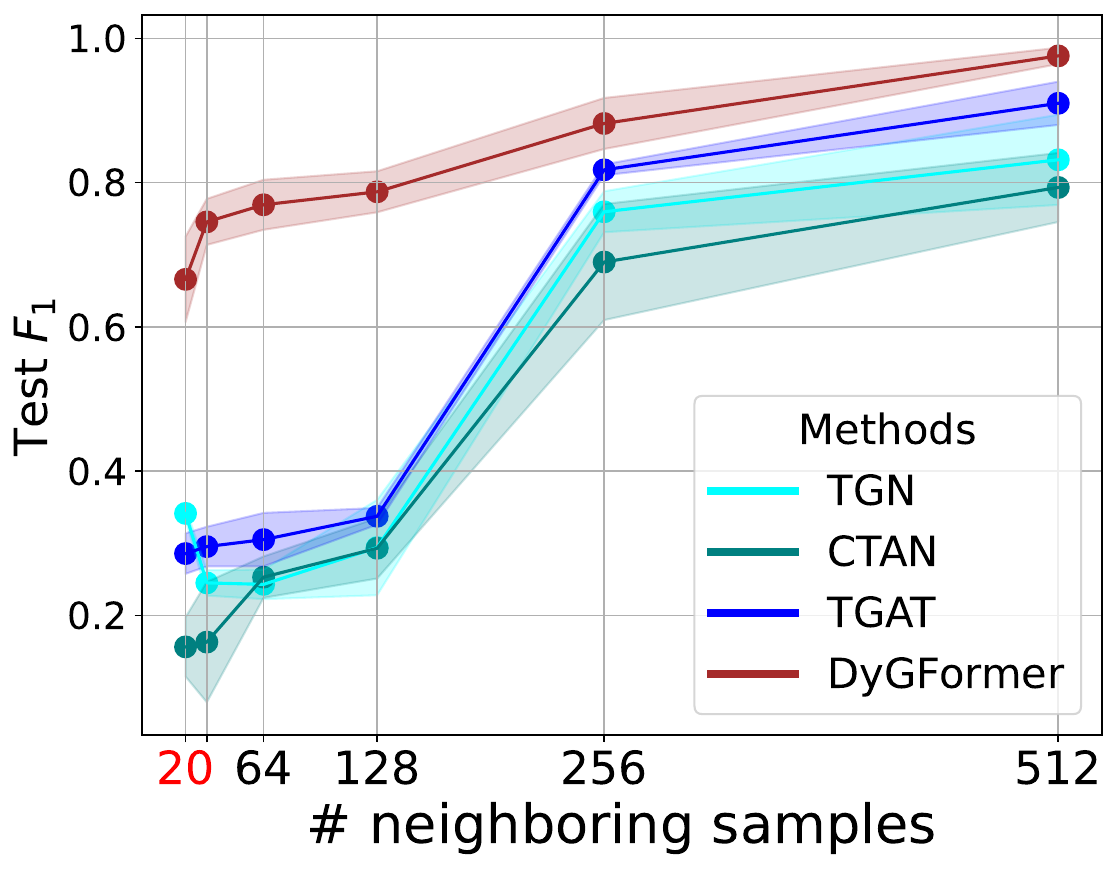}
        \caption{$\Pdet{256, 1}$}
        \label{fig:det_periodic_256_num_nei}
        \label{fig:sub_all_(k, 1)_(0.9, 0.01)_sbm_sto}
    \end{subfigure}
    \hfill
    \begin{subfigure}[b]{0.25\textwidth}
    \centering
    \includegraphics[width=\linewidth]{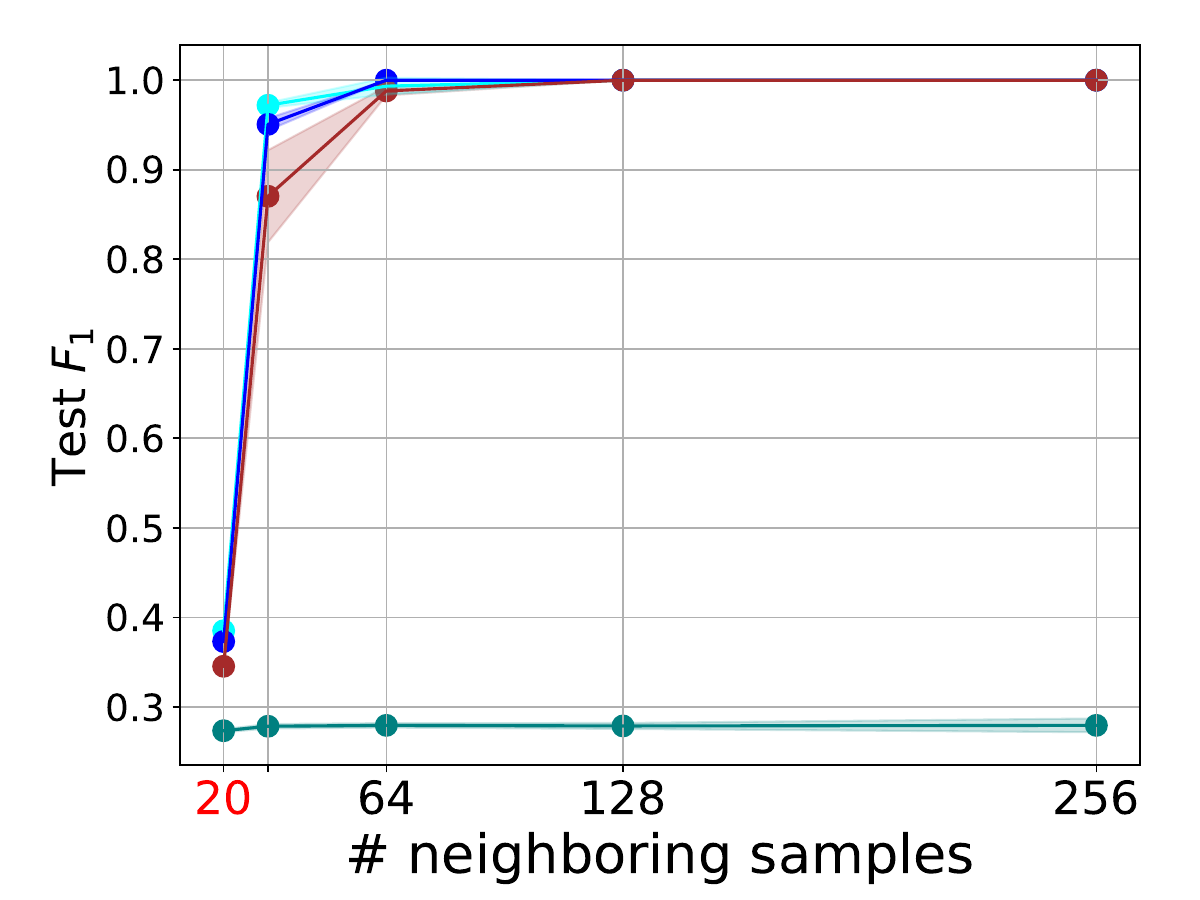}
    \caption{$\causeeffect{64}$}
    \label{fig:cause_effect_64_num_nei}
    \end{subfigure}
    \begin{subfigure}[b]{0.25\textwidth}
    \centering
    \includegraphics[width=\linewidth]{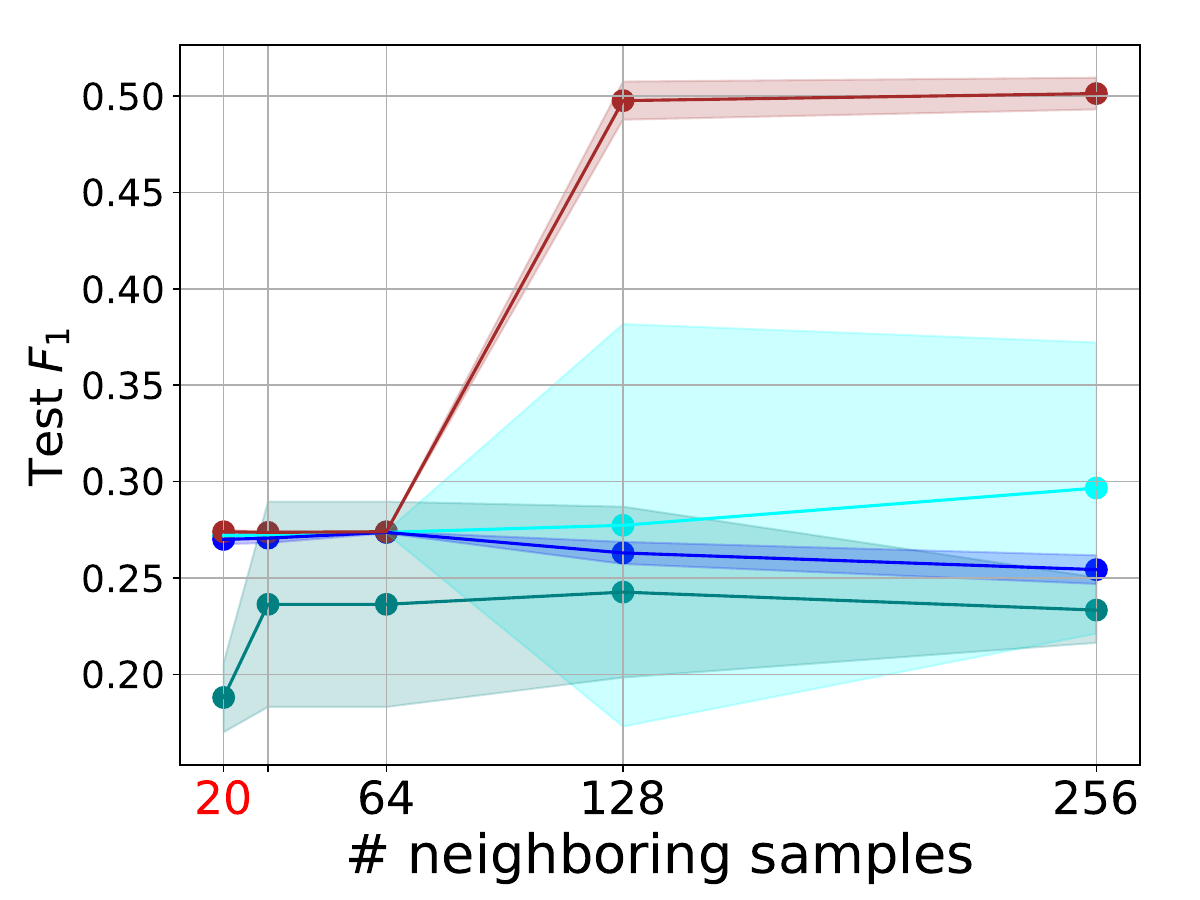}
    \caption{$\causeeffect{256}$}
    \label{fig:cause_effect_256_num_nei}
    \end{subfigure}
    
    \caption{Effect of the number of neighbors on periodic (left), and cause-and-effect (right) tasks .}
    \label{fig:multi_seed_neighbors}
\end{figure}

\looseness-1 Figures~\ref{fig:sto_periodic_8_num_nei} and~\ref{fig:det_periodic_256_num_nei} illustrate performance on periodic tasks as a function of neighbor count. Our analysis reveals that CTAN, TGN, and TGAT exhibit pronounced performance improvements with increased neighbor sampling across both stochastic and deterministic settings, demonstrating their substantial sensitivity to this parameter. For DyGFormer, this parameter directly determines the temporal context length for sequence construction, and it maintains relatively consistent performance, suggesting a more robust architecture with respect to neighbor sampling. Additional analyses are available in Appendix~\ref{apendix:neighbours}. Extending our investigation to cause-and-effect tasks (Figures~\ref{fig:cause_effect_64_num_nei} and~\ref{fig:cause_effect_256_num_nei}), we observe that for $\causeeffect{64}$, TGAT, TGN, and DyGFormer continue to benefit from sampling beyond 32 neighbors, while CTAN plateaus, indicating a fundamental limitation in its capacity and design. For the more challenging $\causeeffect{256}$ task, DyGFormer demonstrates continued improvement with larger neighbor counts (particularly beyond 128), while other methods reach performance saturation. These findings suggest potential for further performance gains through even larger sampling windows (e.g., 512 neighbors), albeit with corresponding computational overhead.

%% file: tex/085conclusion.tex
\section{Conclusion}

We introduced \benchmarkname, a synthetic benchmark designed to systematically probe the temporal reasoning abilities of TGNNs. Our results reveal that no single method excels across all tasks, with performance varying notably across tasks. Attention and memory-based architectures like DyGFormer, TGAT, and TGN show strong performance on long-range and causal tasks, while simpler recurrent models like GC-LSTM excel in periodic tasks. Our findings highlight limitations of current models and the need for architectures tailored to diverse temporal reasoning challenges.

%% file: tex/100appendix.tex
\section{Limitations} \label{appendix:limit}

In this work, our proposed \benchmarkname benchmark probes the capabilities of TGNNs in reasoning over time. One potential limitation is that there might be other capabilities that would be important to test in the future. For example, another challenge in temporal graphs is how to detect anomalies over time. In this work, we focus on temporal link prediction task while the anomaly detection task is often formulated as an edge or node classification task. Currently, \benchmarkname targets the transductive setting where all nodes in the graph are known and we show that current methods already struggle in transductive setting. One future extension is to add the inductive setting for the tasks in \benchmarkname where new nodes might be introduced in later timestamps. Our goal in \benchmarkname is to lay the foundation for systematic evaluation of fundamental capabilities of TGNNs in a controlled environment and we plan to incorporate community feedback and add to the repository of tasks in \benchmarkname in the future.

\section{Broader Impact} \label{appendix:impact}

\textbf{Impact on Temporal Graph Learning.}
It is shown that synthetic, controlled benchmarks such as~\citep{clevr,babi,bsuite} accelerates research in their respective fields. By providing the first synthetic dataset to systematically diagnose the capabilities of TGNNs to reason over time, we expect that \benchmarkname will accelerate the development and evaluation of novel approaches for temporal graph learning. Researchers can experiment on a common benchmark, leading to rapid, robust, and standardized evaluation in this field. In addition, the insights gained from \benchmarkname can inspire novel method or paradigm designs for more robust temporal graph methods. 

\textbf{Potential Negative Impact.} 
\benchmarkname is a robust benchmark for testing TGNN's capabilities. However, we also acknowledge that this is a synthetic benchmark thus we advise researchers to utilize both \benchmarkname and real world temporal networks to obtain a more comprehensive understanding of model performance in both controlled and real-world experiments. As a synthetic benchmark, it is possible that certain real-world dynamics and interactions are not fully captured by the benchmark. Thus, if research methods are tested solely on \benchmarkname, the results might not directly translate to certain real-world networks. Therefore, we advise the community to first test on \benchmarkname benchmark and then also apply their model on real world benchmarks. We also plan to update the benchmark regularly to incorporate community feedback.


\section{Dataset and Code Licenses and Links} \label{appendix:licenses}
The code and datasets can be accessed at \url{https://github.com/alirezadizaji/T-GRAB} and \url{https://huggingface.co/datasets/Gilestel/T-GRAB}, respectively. Both are released under the Apache 2.0 license.

\section{Experimental Details}\label{appendix:experimental-details}
In this section, we provide additional details on our experiments.
\subsection{Dataset split}
Each dataset is chronologically divided into training, validation, and test splits. For periodicity tasks, it’s essential to ensure that the data splits always preserve complete periods. In our experiments on periodicity tasks, we allocated 40, 4, and 4 full periods for the training, validation, and test sets, respectively. For other tasks, we divided the data by distributing 80\%, 10\%, and 10\% of all snapshots into the training, validation, and test sets.

\subsection{Method training details}

\looseness-1 In this section, we describe how we trained our models. At every epoch, models are trained on the training set and evaluated on the validation set. We used a single layer of message passing in all of our experiments. As outlined in \Cref{sec:discussion}, whenever applicable and unless mentioned otherwise, we perform a most-recent neighbourhood sampling during temporal message passing with size 20. 

Whenever a model reaches its best validation performance, it is subsequently evaluated on the test set. Reported results are averaged over five different random seeds. Each model was trained with a patience of 100 epochs, where training would stop if the training loss failed to improve for 100 consecutive epochs, without enforcing a fixed maximum number of epochs. We used the Adam optimizer with a learning rate of $10^{-4}$ and a batch size of one snapshot, meaning that the $i$-th batch includes all links from timestep $i$ (see Algorithm~\ref{alg:train}). As per standard practice, to predict the probability of a link between $i$ and $j$, we concatenate their hidden representations $h_i$ and $h_j$ before passing it to a two-layer MLP to predict a probability score between 0 and 1. 
\begin{algorithm}
    \KwInput{Number of training epochs $N$, dynamic graph model \textit{model}, early stopping patience $P$ based on training loss}
    \KwParam{Final timesteps for training $T^{\text{(train)}}$, validation $T^{\text{(val)}}$, and testing $T$, best observed training loss $L^{\text{(best)}}$ and its epoch $e^{\text{(best)}}$, best validation metric $F^{\text{(best)}}_1$.}
    \caption{Discrete-Time Model Training and Evaluation}
    \label{alg:train}

    $F^{\text{(best)}}_1 \leftarrow 0$\;
    $L^{\text{(best)}} \leftarrow \infty$, $e^{\text{(best)}} \leftarrow 0$\;

    \For{$e$ in $\{1, \dots, N\}$ }{
        model.reset\_mem()\Comment*[r]{Clear model's internal memory (if applicable)}
        Initialize empty list $L_s$\;

        \For{$t$ in $\{1, \dots, T^{\text{(train)}}\}$\Comment*[r]{\textcolor{blue}{Training loop}}}{
            $X \leftarrow$ model.get\_emb($G_t$)\Comment*[r]{Extract embeddings for snapshot $G_t$}
            $L \leftarrow$ compute\_loss($X$, $G_t$)\Comment*[r]{Calculate loss for current snapshot}
            Append $L$ to $L_s$\;
            model.backprop($L$)\Comment*[r]{Update parameters via backpropagation}
            model.update($G_t$)\Comment*[r]{Perform internal updates (e.g. memory state)}
        }

        Compute average training loss $L_a$ from $L_s$\;

        \If{$L_a < L^{\text{(best)}}$}{
            $L^{\text{(best)}} \leftarrow L_a$\;
            $e^{\text{(best)}} \leftarrow e$\;
        }

        \If{$e - e^{\text{(best)}} \geq P$\Comment*[r]{Early stopping check}}{
            \textbf{exit}\;
        }

        Initialize empty list $F^{\text{(val)}}_1$\;
        \For{$t$ in $\{T^{\text{(train)}}+1, \dots, T^{\text{(val)}}\}$ \Comment*[r]{\textcolor{blue}{Validation loop}}}{
            $X \leftarrow$ model.get\_emb($G_t$)\;
            $F_1 \leftarrow$ compute\_metric($X$, $G_t$)\;
            Append $F_1$ to $F^{\text{(val)}}_1$\;
            model.update($G_t$)\;
        }
        Compute average validation metric $F_1$ from $F^{\text{(val)}}_1$\;

        \If{$F_1 > F^{\text{(best)}}_1$ \Comment*[r]{\textcolor{blue}{Testing loop, if validation improves}}}{
            Save current model state\;
            $F^{\text{(best)}}_1 \leftarrow F_1$\;
            Initialize empty list $F^{\text{(test)}}_1$\;
            \For{$t$ in $\{T^{\text{(val)}}+1, \dots, T\}$}{
                $X \leftarrow$ model.get\_emb($G_t$)\;
                $F_1 \leftarrow$ compute\_metric($X$, $G_t$)\;
                Append $F_1$ to $F^{\text{(test)}}_1$\;
                model.update($G_t$)\;
            }
            Compute final test metric $F^{\text{(ret)}}_1$ as the average of $F^{\text{(test)}}_1$\;
        }
    }
    \Return $F^{\text{(ret)}}_1$\;
\end{algorithm}

In our experiments with TGAT and DyGFormer, we used the implementation provided by DyGLib \cite{dygformer}. For EvolveGCN, we used the original implementation of EvolveGCN-O found at \cite{evolvegcn}. For GAT and GCN, we used PyTorch Geometric's \footnote[2]{\url{https://pytorch-geometric.readthedocs.io/en/latest/}} implementations, and PyTorch Geometric-Temporal implementations for T-GCN and GC-LSTM \footnote[3]{\url{https://pytorch-geometric-temporal.readthedocs.io/en/latest/modules/root.html}}. For CTAN and TGN, however, we made a few slight adjustments:

\paragraph{CTAN.} In our initial experiments with CTAN, the original implementation of CTAN \cite{ctan} failed during training, producing large values that eventually led to NaNs. The issue originated from a specific MLP layer applied to the node embeddings from the memory module, which lacked both an activation function and batch normalization. To address this and enable stable training on our tasks, we introduced an activation layer after this MLP, which provided training stability.
\paragraph{TGN.}
Our experiments with TGN used TGB's \cite{tgb} implementation of TGN, which matches that of PyTorch Geometric. However, in our early experiments, we observed that the default choice of \texttt{TransformerConv} as the embedding layer performed substantially worse than TGAT. Hence, we've replaced \texttt{TransformerConv} with TGAT in all our experiments, while keeping all other default architectural components unchanged.

\section{Additional Dataset Details}
In this section, we outline the dataset sizes for each task, providing additional salient details where appropriate. 

\subsection{Deterministic Periodicity Tasks}\label{appendix:periodicity-deterministic-stats}
Table~\ref{tab:det_periodic_datasets} presents detailed statistics for our deterministic periodicity tasks. Node features were represented using one-hot encoding of node IDs, while all edge features were set to a constant value of one. For $\Pdet{2, n}$ and $\Pdet{k, 1}$, we generated eight variations to adjust the difficulty of counting and memorization.
\begin{table}[h]
\caption{Deterministic periodicity task statistics}
\centering
\begin{tabular}{cc | cc | c}
\toprule
\multicolumn{2}{c}{Counting Periodicity} & \multicolumn{2}{c}{Memorizing Periodicity} & \\
\cmidrule(lr){1-2} \cmidrule(lr){3-4}
\textbf{Dataset} & \textbf{$\#$ Edges } & \textbf{Dataset} & \textbf{$\#$ Edges } & \textbf{$\#$ timesteps}  \\
\midrule
$\Pdet{2, 1}$ & $10,560$ & $\Pdet{2, 1}$ & $10,560$ & $96$  \\[0.24em]
$\Pdet{2, 2}$ & $21,120$ & $\Pdet{4, 1}$ & $19,584$ & $192$ \\[0.24em]
$\Pdet{2, 4}$ & $42,240$ & $\Pdet{8, 1}$ & $37,440$ & $384$ \\[0.24em]
$\Pdet{2, 8}$ & $84,480$ & $\Pdet{16, 1}$ & $78,720$ & $768$ \\[0.24em]
$\Pdet{2, 16}$ & $168,960$ & $\Pdet{32, 1}$ & $158,688$ & $1,536$ \\[0.24em]
$\Pdet{2, 32}$ & $337,920$ & $\Pdet{64, 1}$ & $306,336$ & $3,072$ \\[0.24em]
$\Pdet{2, 64}$ & $675,840$ & $\Pdet{128, 1}$ & $602,976$ & $6,144$ \\[0.24em]
$\Pdet{2, 128}$ & $1,351,680$ & $\Pdet{256, 1}$ & $1,206,336$ & $12,288$ \\[0.24em]
\bottomrule
\end{tabular}
\vspace{1em}
\label{tab:det_periodic_datasets}
\end{table}

\subsection{Stochastic Periodicity Tasks}\label{appendix:periodicity_stochastic_design}
In the stochastic tasks $\Psto{k,1}$, there are $k$ SBMs distributions $D_1,\cdots, D_k$ that all share the same inter-community and intra-community edge probabilities but have different community structures. For each $i=1,\dots,k$, the community structure for $D_i$ is randomly generated by partitioning the 100 nodes in 3 random subsets of equal sizes~(33-33-34).  At time step $t$, the static $G_t$ is drawn from the distribution $D_{(t \text{ mod } k) +1}$. 
For all experiments, the inter-community edge probability is equal to 0.01 while we explore two settings for the intra-community edge probability: an easy one $p=0.9$ and a more challenging one $p=0.5$.The number of edges varies between 219,960 and 9,144,440, depending on the period length $k$ and the intra-probability $p$.

\subsection{Cause \& Effect Tasks}
Algorithm~\ref{alg:cause_effect} provides the pseudocode for generating the cause-and-effect tasks, while Table~\ref{tab:cause_and_effect_stats} summarizes their statistics. Each task is constructed to include exactly 4000 effect subgraphs, resulting in a total of \( 4000 + \ell \) timesteps per task.

\begin{algorithm}[hbt!]
\caption{Cause \& Effect Task Generation}\label{alg:two}
\label{alg:cause_effect}
\KwInput{Erdős-Rényi model \textit{ER} with number of nodes \textbf{$n^{\text{(er)}}$} and edge probability \textbf{$p^{\text{(er)}}$}, temporal lag \textbf{$\ell$} between cause and effect subgraphs, total number of timesteps $T$.}

\KwParam{Node set $V$ shared at each timestep $t$, cause subgraphs $G^{(\text{c})}_t$, effect subgraphs $G^{(\text{e})}_t$, and the set of active nodes $N_t$ at each timestep.}

Initialize an empty list $\mathcal{G}$\;

\For{$t \in \{1, \dots, \ell\}$ \Comment*[r]{First $\ell$ steps contain only cause subgraphs}} {
    Generate cause subgraph: $G^{(\text{c})}_t = \text{ER}(n^{\text{(er)}}, p^{\text{(er)}})$\;
    Append $G^{(\text{c})}_t$ to $\mathcal{G}$\;
}

\For{$t \in \{\ell + 1, \dots, T\}$} {
    Generate cause subgraph: $G^{(\text{c})}_t = \text{ER}(n^{\text{(er)}}, p^{\text{(er)}})$\;
    
    Create effect edges: $E^{(\text{e})}_t = \{(0, m) \mid m \in N_{t-\ell}\}$\;
    
    Build effect subgraph: $G^{(\text{e})}_t = (V, E^{(\text{e})}_t)$\;
    
    Merge cause and effect subgraphs: $G_t = G^{(\text{e})}_t + G^{(\text{c})}_t$\;
    
    Append $G_t$ to $\mathcal{G}$\;
}
\Return $\mathcal{G}$
\end{algorithm}

\begin{table}[h]
    \caption{Cause \& Effect task statistics}
    \centering
    \begin{tabular}{c|c|c}
        \toprule
        Dataset & $\#$ Edges & $\#$ timesteps \\
        \midrule
        $\causeeffect{1}$& 164,856 & 4,001 \\ 
        $\causeeffect{4}$& 165,142 & 4,004 \\ 
        $\causeeffect{16}$& 165,488 & 4,016 \\ 
        $\causeeffect{64}$& 167,182 & 4,064 \\ 
        $\causeeffect{256}$& 174,470 & 4,256 \\ 
    \end{tabular}
    \label{tab:cause_and_effect_stats}
\end{table}

\subsection{Spatio-Temporal Long Range Tasks}
Table~\ref{tab:spatio_long_range_stats} summarizes the statistics of the spatio-temporal long-range datasets, where the number of edges is determined by the temporal parameter~\( \ell \) and the spatial distance parameter~\( d \). The task generation process for spatio-temporal long-range follows a similar approach to the cause-and-effect setup described in Algorithm~\ref{alg:cause_effect}, with one key difference: in this case, the effect edges of~\( G_t \) connect the target node to the last three nodes encountered along each path in~\( G_{t - \ell} \).

\begin{table}[h]
    \caption{Spatio-temporal Long Range task statistics}
    \centering
    \begin{tabular}{c|ccccc|c}
        \toprule
        \multirow{2}{*}{Dataset} & \multicolumn{5}{c|}{$\#$ Edges} & \multirow{2}{*}{$\#$ Timesteps} \\
        & $d=1$ & $d=2$ & $d=4$ & $d=8$ & $d=16$ & \\
        \midrule
        $\longrange{\ell=1, d}$ & 48,006 & 72,012 & 120,024 & 216,048 & 408,096 & 4,001 \\
        $\longrange{\ell=4, d}$ & 48,024 & 72,048 & 120,096 & 216,192 & 408,384 & 4,004 \\
        $\longrange{\ell=16, d}$ & 48,096 & 72,192 & 120,384 & 216,768 & 409,536 & 4,016 \\
        $\longrange{\ell=32, d}$ & 48,192 & 72,384 & 120,768 & 217,536 & 411,072 & 4,032 \\
    \end{tabular}
    \label{tab:spatio_long_range_stats}
\end{table}

\section{Model Sizes}\label{appendix:parameters}
Table~\ref{tab:model_num_params} lists the number of parameters for each model trained on our set of tasks. Interestingly, despite DyGFormer having the highest parameter count among all models, it fails to deliver the best performance, as shown in Table~\ref{tab:overall}. In fact, across all tasks, other methods consistently outperform DyGFormer, indicating that a larger model size does not necessarily lead to superior results in every setting.
\begin{table}[h]

    \caption{Number of parameters per model}
    \label{tab:model_num_params}
    \centering
    \begin{tabular}{c|c}
        Model & $\#$ parameters ($\times 10^3$) \\
        \toprule
        CTAN & 165 \\
        DyGFormer & 606 \\
        TGAT & 234 \\
        TGN & 207 \\
        \midrule
        GC-LSTM & 233 \\
        EGCN & 163 \\
        T-GCN & 187 \\
        GCN & 63 \\
        GAT & 63 \\
        \bottomrule
    \end{tabular}
\end{table}

\section{Compute Resources} \label{appendix:compute}
All experiments were conducted on a single NVIDIA GPU, utilizing one of the following models: RTX 8000, L40S (48GB), A100 (40GB), or V100 (32GB). Runtimes varied based on task complexity and model architecture, ranging from approximately 5 minutes to 4 days.
 
\section{Additional Analysis}

\subsection{Counting and Periodicity}\label{appendix:pertimesteps_counting}
\begin{figure}[H]
    \centering
    \begin{subfigure}[b]{0.44\textwidth}
        \centering
        \includegraphics[width=\linewidth]{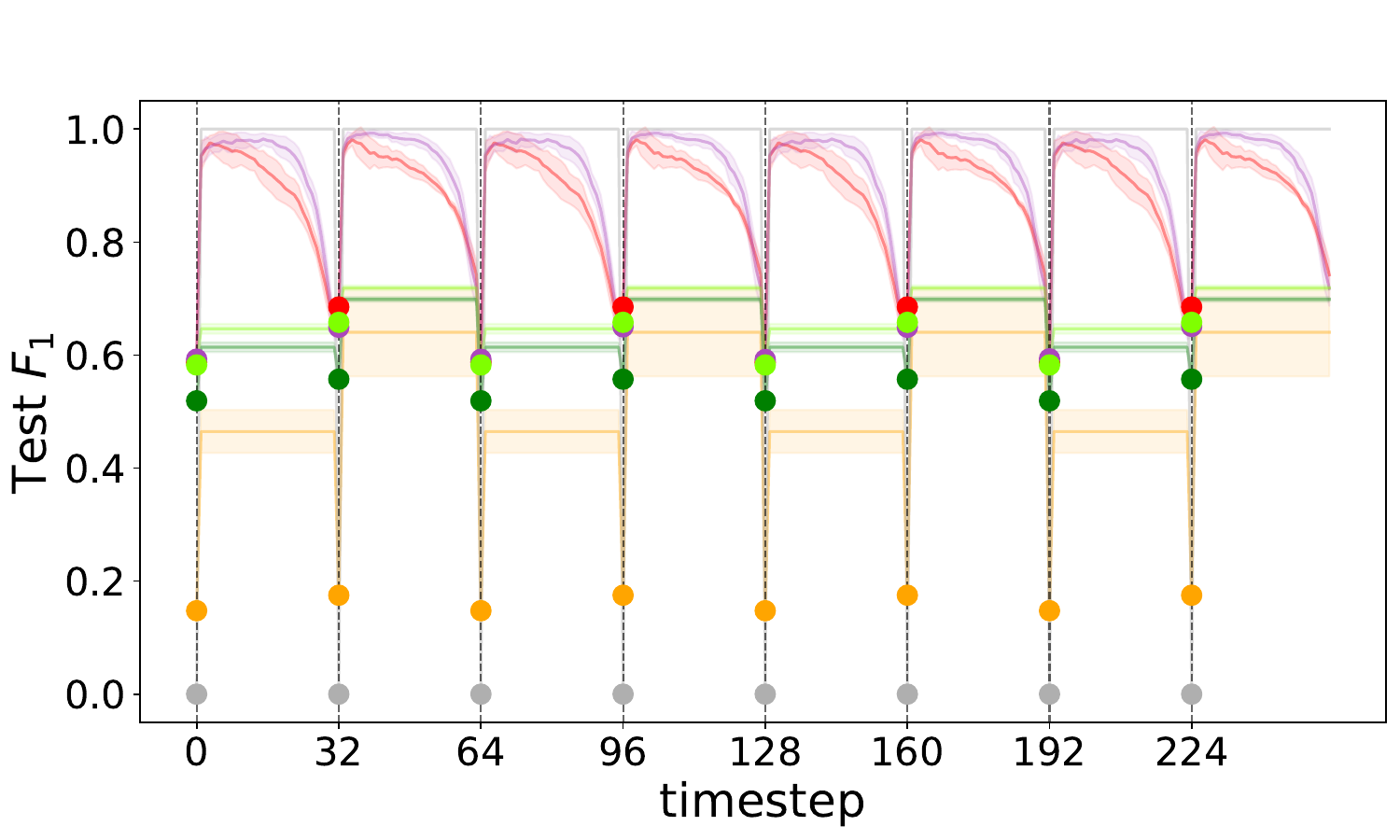}
        \caption{DTDG methods}
        \label{fig:sub_dtdg_det_(32,1)_per_timestep}
    \end{subfigure}
    \begin{subfigure}[b]{0.44\textwidth}
        \centering
        \includegraphics[width=\linewidth]{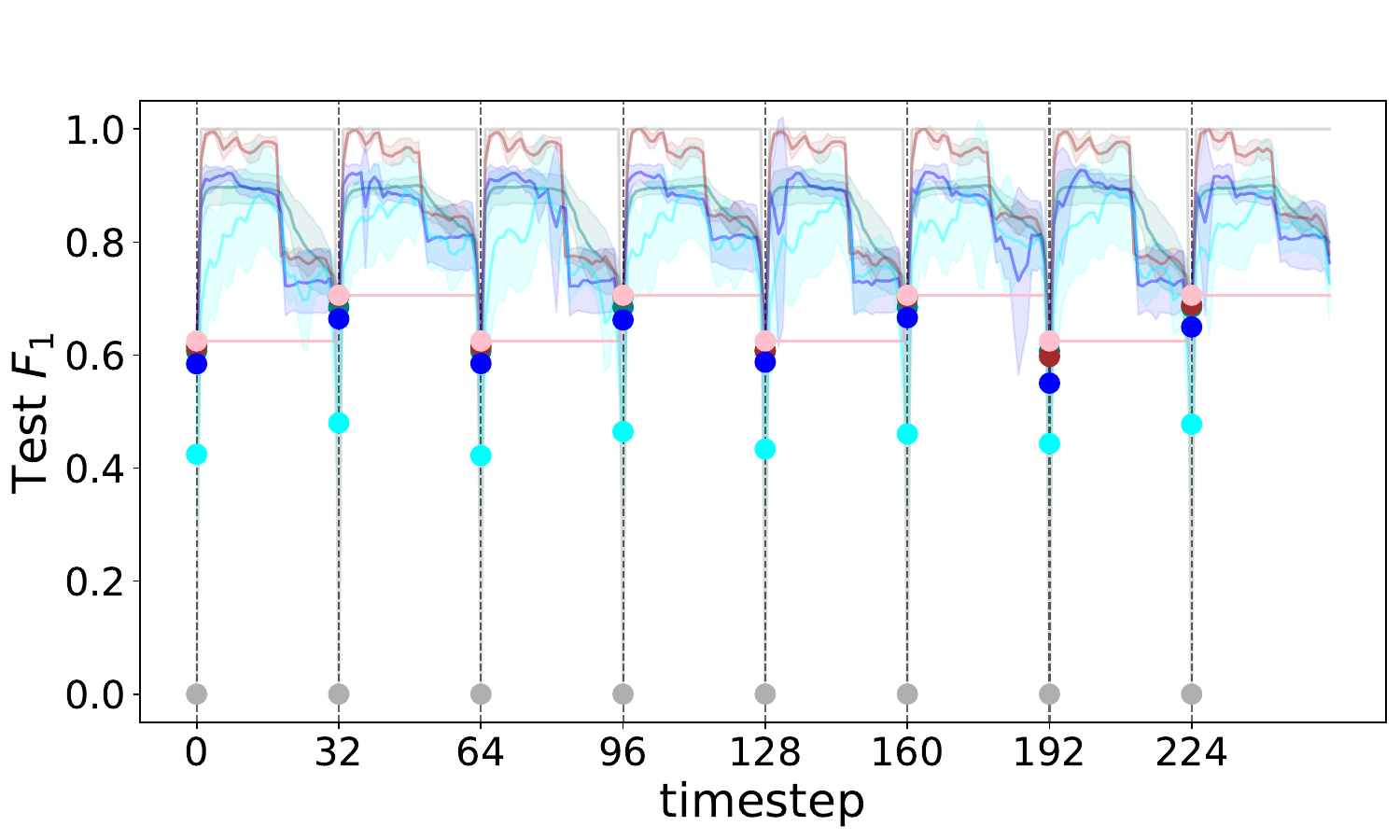}
        \caption{CTDG methods}
    \label{fig:sub_ctdg_det_(32,1)_per_timestep}
    \end{subfigure}
    \begin{subfigure}[b]{0.1\textwidth}
        \centering
        \raisebox{12mm}{
        \includegraphics[width=\linewidth]{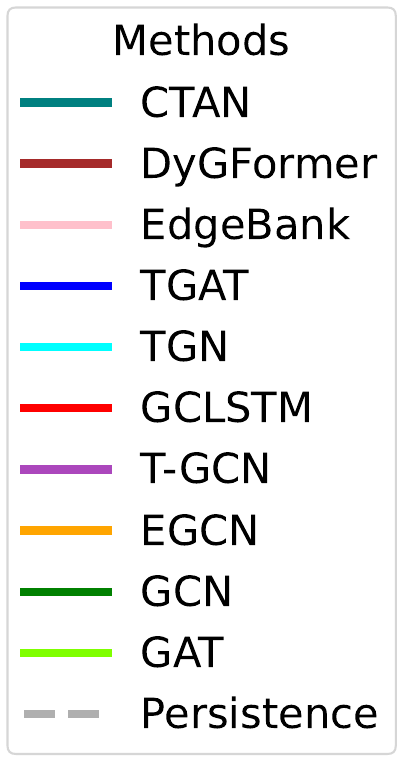}}
    \end{subfigure}
    
    \caption{The $F_1$ score at each timestep over the evaluation period is shown for DTDG (left) and CTDG (right) methods on the $\Pdet{2, 32}$ dataset.}
    \label{fig:(2,N)_per_timestep}
\end{figure}
 Figure~\ref{fig:(2,N)_per_timestep} offers a detailed, timestep-level view of model performance throughout the evaluation period. It shows that most models tend to lose performance at the start of each repetition phase and struggle to transition cleanly to the next pattern at the predefined change points (indicated by vertical lines). A notable observation is the sharp increase in performance immediately after these change points among temporal graph learning methods. This implies that these models rely on seeing the new snapshot once before adapting their predictions, which is indicative of them learning a function reminiscent of the persistence heuristic rather than properly counting. However, we note that the functions that the models learn are not \textit{exactly} equal to the persistence heuristic, as the $F_1$ score decreases between changepoints. 

\subsection{Impact of historical neighbour sampling.} \label{apendix:neighbours}
Figure~\ref{fig:(K,1)_num_neighbors_appendix} presents plots analogous to those in Figure~\ref{fig:multi_seed_neighbors}, but for additional values of $k$ in the periodicity tasks. These results offer additional insights into the impact of varying the number of neighboring samples.

In deterministic scenarios (Figures~\ref{fig:sub_num_neighbors_det_(32,1)} and \ref{fig:sub_num_neighbors_det_(64,1)}), TGAT and DyGFormer consistently exhibit robust performance, demonstrating minimal sensitivity to variations in $k$. Their $F_1$ scores remain largely stable across the tested range of neighbor samples. Conversely, TGN and CTAN display marked sensitivity to this hyperparameter. For instance, increasing k leads to substantial $F_1$ score improvements for both models, with CTAN showing gains of up to approximately 20\% and TGN similarly benefiting, underscoring their reliance on a sufficiently large neighborhood for optimal performance in these noise-free environments.


Under stochastic conditions (Figures~\ref{fig:sub_num_neighbors_sto_(4,1)} and \ref{fig:sub_num_neighbors_sto_(16,1)}), TGAT continues to demonstrate low sensitivity to $k$, maintaining relatively stable $F_1$ scores, akin to its behavior in the deterministic setting. DyGFormer, however, struggles significantly in the presence of noise, yielding considerably lower $F_1$ scores across all values of $k$, which suggests a lack of robustness to stochasticity in the temporal data. TGN's performance in stochastic settings appears somewhat erratic; while it benefits from more neighbors initially in $\Psto{4, 1}$, this trend does not hold consistently, as seen in $\Psto{16, 1}$ where performance degrades significantly at $k = 128$. In contrast, CTAN consistently leverages an increased number of historical neighbors, with its $F_1$ score progressively improving with larger $k$, mirroring its behavior in the deterministic tasks and indicating a capacity to filter noise or extract a clearer signal when more historical context is available.


\begin{figure}[H]
    \centering
    \begin{subfigure}[b]{0.24\textwidth}
        \centering
        \includegraphics[width=\linewidth]{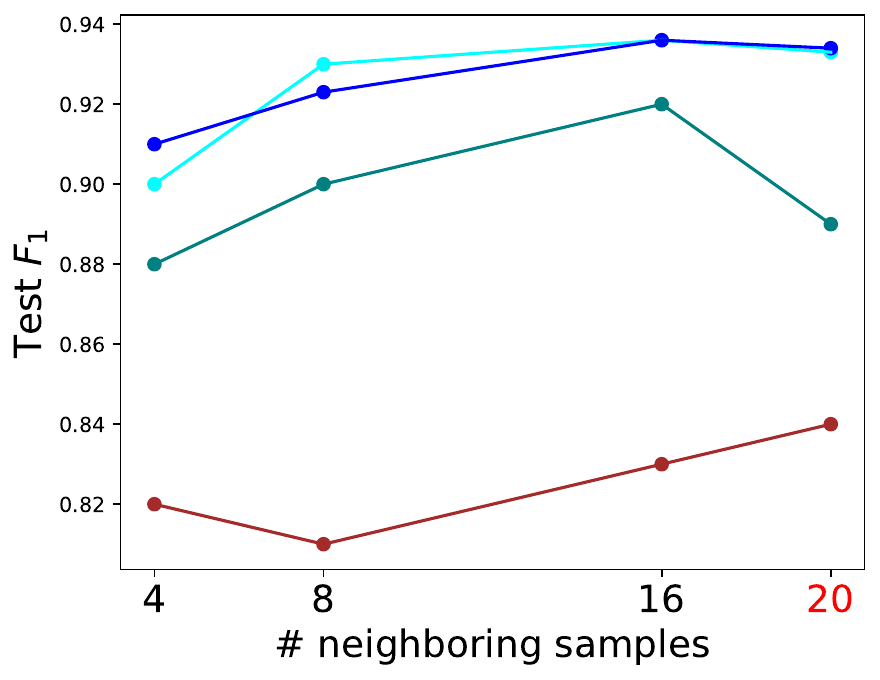}
        \caption{$\Psto{4, 1}$}
        \label{fig:sub_num_neighbors_sto_(4,1)}
    \end{subfigure}
    \begin{subfigure}[b]{0.24\textwidth}
        \centering
        \includegraphics[width=\linewidth]{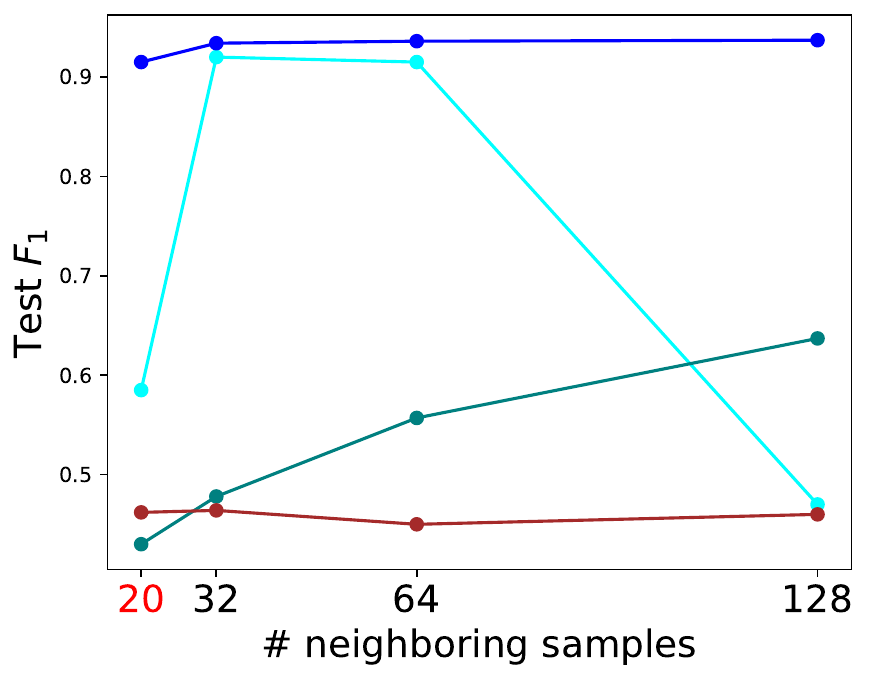}
        \caption{$\Psto{16, 1}$}
        \label{fig:sub_num_neighbors_sto_(16,1)}
    \end{subfigure}
    \begin{subfigure}[b]{0.24\textwidth}
        \centering
        \includegraphics[width=\linewidth]{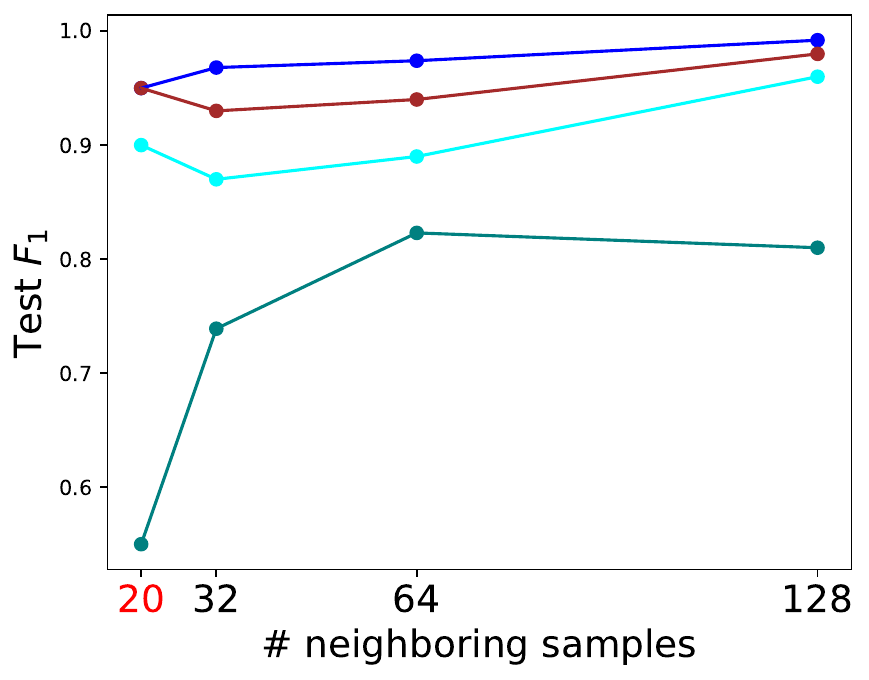}
        \caption{$\Pdet{32, 1}$}
        \label{fig:sub_num_neighbors_det_(32,1)}
    \end{subfigure}
    \begin{subfigure}[b]{0.24\textwidth}
        \centering
        \includegraphics[width=\linewidth]{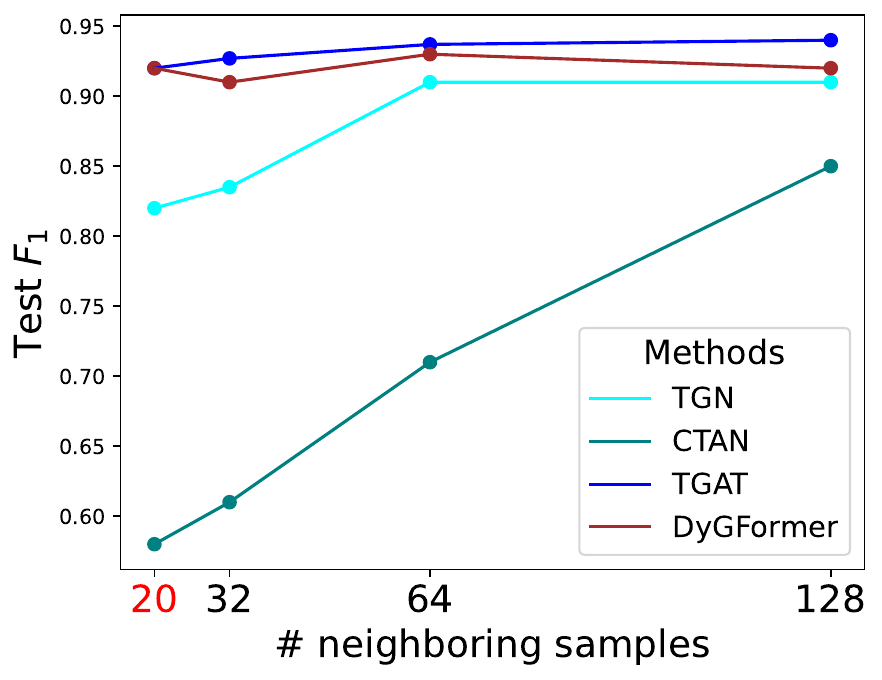}
        \caption{$\Pdet{64, 1}$}
        \label{fig:sub_num_neighbors_det_(64,1)}
    \end{subfigure}
    \caption{The impact of number of historical neighbor sampling on stochastic and deterministic periodic tasks.}
    \label{fig:(K,1)_num_neighbors_appendix}
\end{figure}

\subsection{Spatio-temporal long-range tasks (full methods)} \label{apendix:spatio_temporal_full_methods}
\begin{figure}[H]
    \centering
    \begin{subfigure}[b]{0.22\textwidth}
        \centering
        \includegraphics[width=\linewidth]{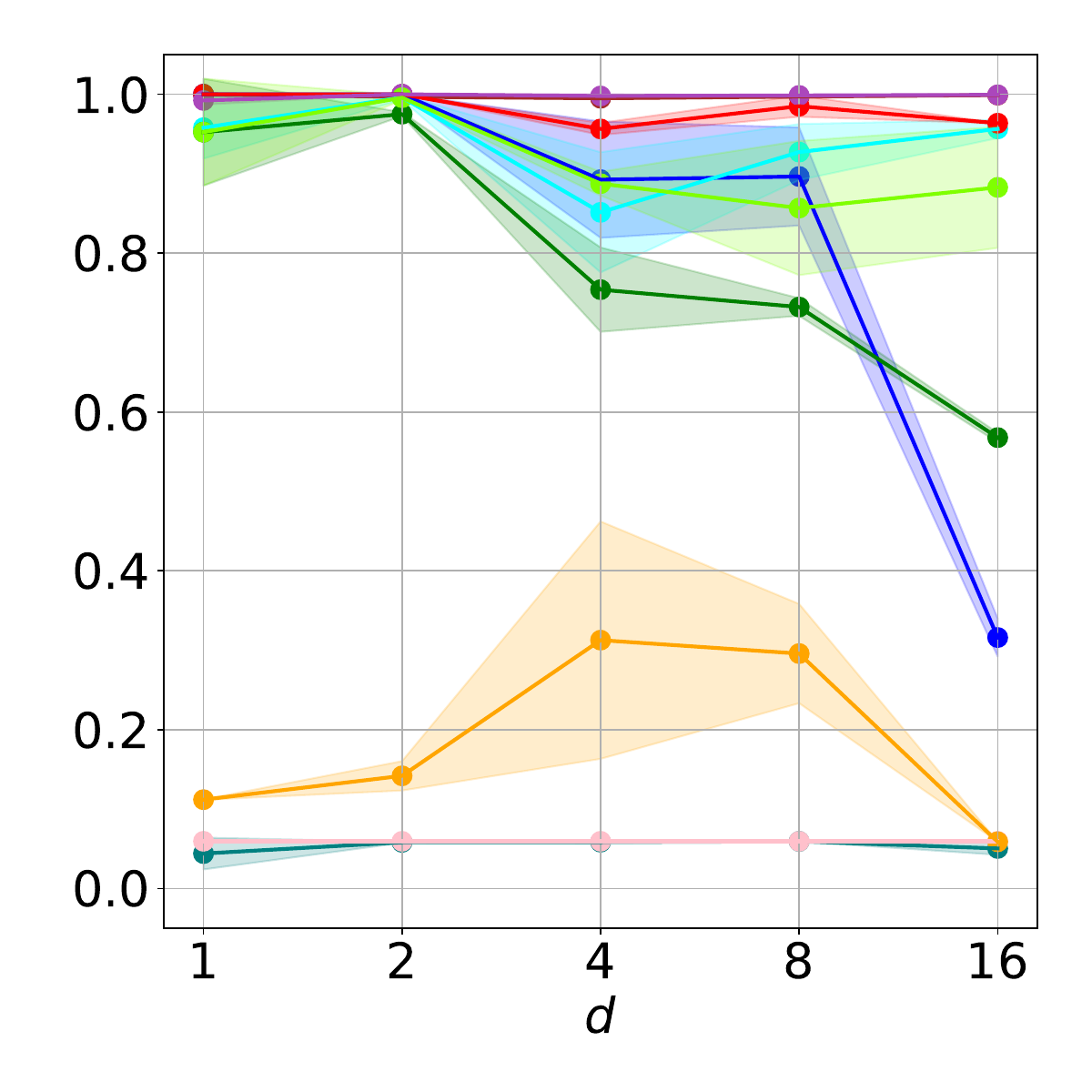}
        \caption{$(\ell=1, d)$}
        \label{fig:sub_long_range_full_(1,d)}
    \end{subfigure}
    \begin{subfigure}[b]{0.22\textwidth}
        \centering
        \includegraphics[width=\linewidth]{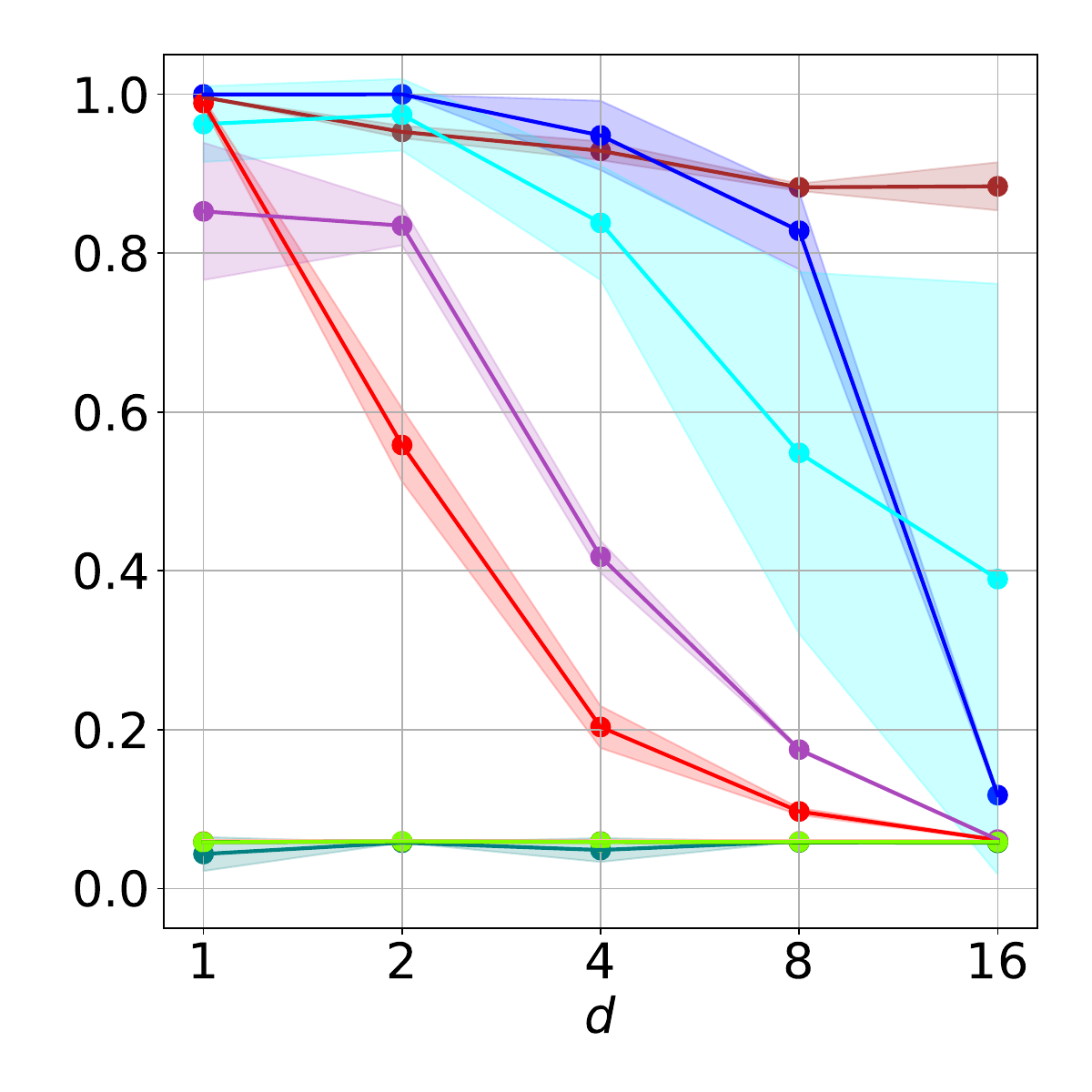}
        \caption{$(\ell=4, d)$}
        \label{fig:sub_long_range_full_(4,d)}
    \end{subfigure}
    \centering
    \begin{subfigure}[b]{0.22\textwidth}
        \centering
        \includegraphics[width=\linewidth]{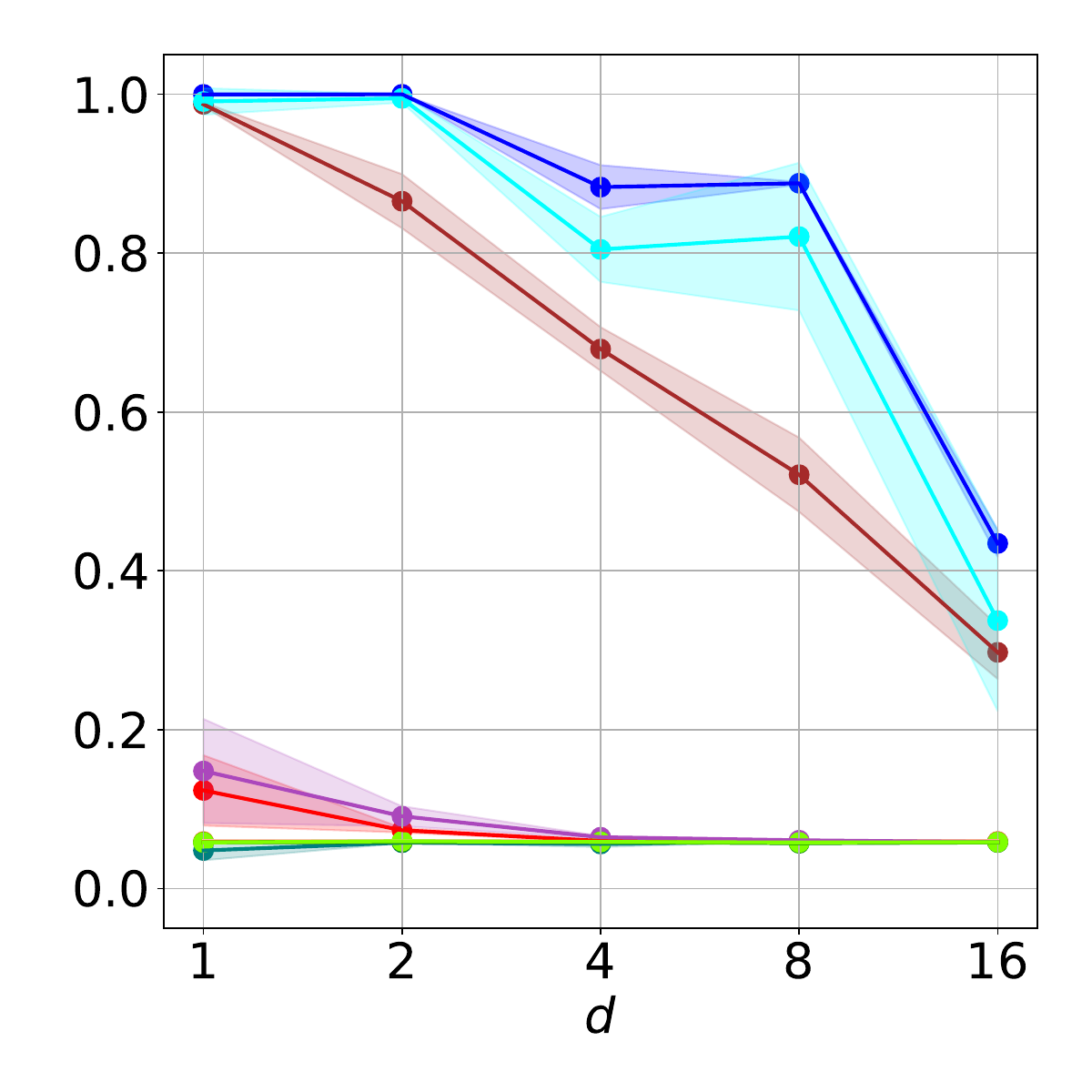}
        \caption{$(\ell=16, d)$}
        \label{fig:sub_long_range_full_(16,d)}
    \end{subfigure}
    \begin{subfigure}[b]{0.22\textwidth}
        \centering
        \includegraphics[width=\linewidth]{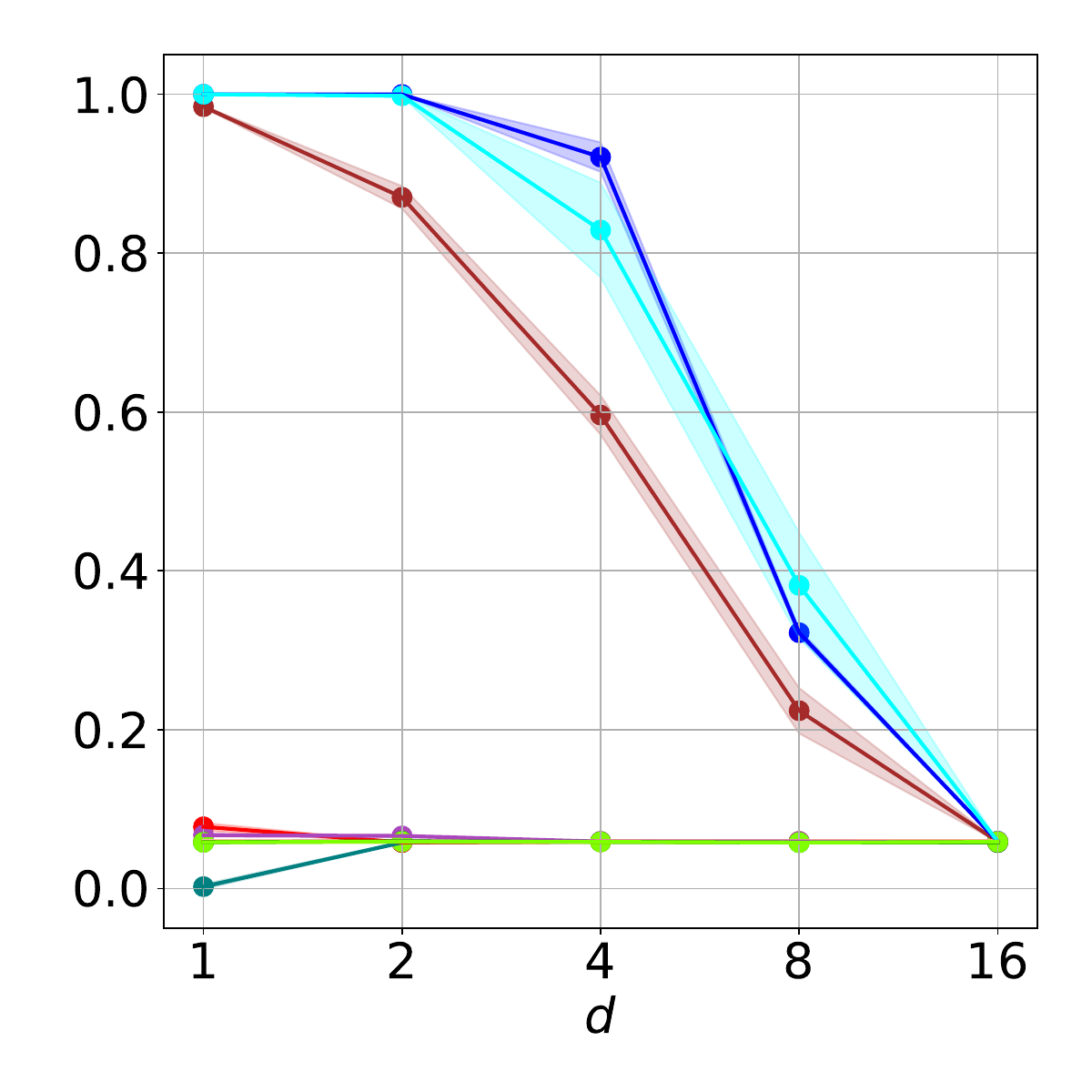}
        \caption{$(\ell=32, d)$}
        \label{fig:sub_long_range_full_(32,d)}
    \end{subfigure}
    \begin{subfigure}[b]{0.08\textwidth}
        \centering
        \raisebox{15mm}{
        \includegraphics[width=\linewidth]{figs/legends/CTDG_DTDG_all_w_edgebank_wo_persistence.pdf}}
    \end{subfigure}
    
    \caption{(Full methods version) Long-range spatio-temporal tasks across varying spatial and temporal distances.}
    \label{fig:plots_long_range_full}
\end{figure}

Figure~\ref{fig:plots_long_range_full} presents the evaluation results on spatio-temporal long-range tasks using the full set of methods, rather than the subset of top-performing models shown in \Cref{fig:plots_long_range}. In this analysis, we additionally include CTAN, EvolveGCN, GCN, and GAT. Interestingly, CTAN performs poorly across all tasks, despite its explicit design to capture long-range dependencies. GAT, on the other hand, performs reasonably well when the temporal dependency is small ($\ell=1$), suggesting that static graphs can sufficiently model simple temporal patterns. However, its performance declines sharply as the temporal dependency increases, resulting in a failure to capture both higher-order temporal and spatial dependencies.

